\documentclass[10pt,twocolumn,letterpaper]{article}

\usepackage[pagenumbers]{cvpr} %

\usepackage[dvipsnames]{xcolor}

\newcommand{\prompt}[1]{\small\textit{``#1''}}

\usepackage{algorithm}
\usepackage[noend]{algpseudocode}
\usepackage{float}
\usepackage{amsmath}
\usepackage{amsfonts}

\usepackage[utf8]{inputenc} %
\usepackage[T1]{fontenc}    %
\usepackage{url}            %
\usepackage{booktabs}       %
\usepackage{amsfonts}       %
\usepackage{nicefrac}       %
\usepackage{microtype}      %
\usepackage{xcolor}         %
\usepackage{graphicx}

\usepackage{array}
\usepackage{multirow}
\usepackage{makecell}
\usepackage[most]{tcolorbox}

\definecolor{cvprblue}{rgb}{0.21,0.49,0.74}
\usepackage[pagebackref,breaklinks,colorlinks,citecolor=cvprblue]{hyperref}

\newcommand{\method}{AutoVFX}
\newcommand{\redcross}{\textcolor{red}{\texttimes}}
\newcommand{\greenmark}{\textcolor{green}{\checkmark}}

\title{\method: Physically Realistic Video Editing from Natural Language Instructions}

\author{
    Hao-Yu Hsu \quad Zhi-Hao Lin \quad Albert J. Zhai \quad Hongchi Xia \quad Shenlong Wang\\
    University of Illinois at Urbana-Champaign \\
    \href{https://haoyuhsu.github.io/autovfx-website/}{https://haoyuhsu.github.io/autovfx-website/}
}

\begin{document}

\twocolumn[{
\renewcommand\twocolumn[1][]{#1}
\maketitle

\begin{center}
    \vspace{-8mm}
    \includegraphics[width=\textwidth]{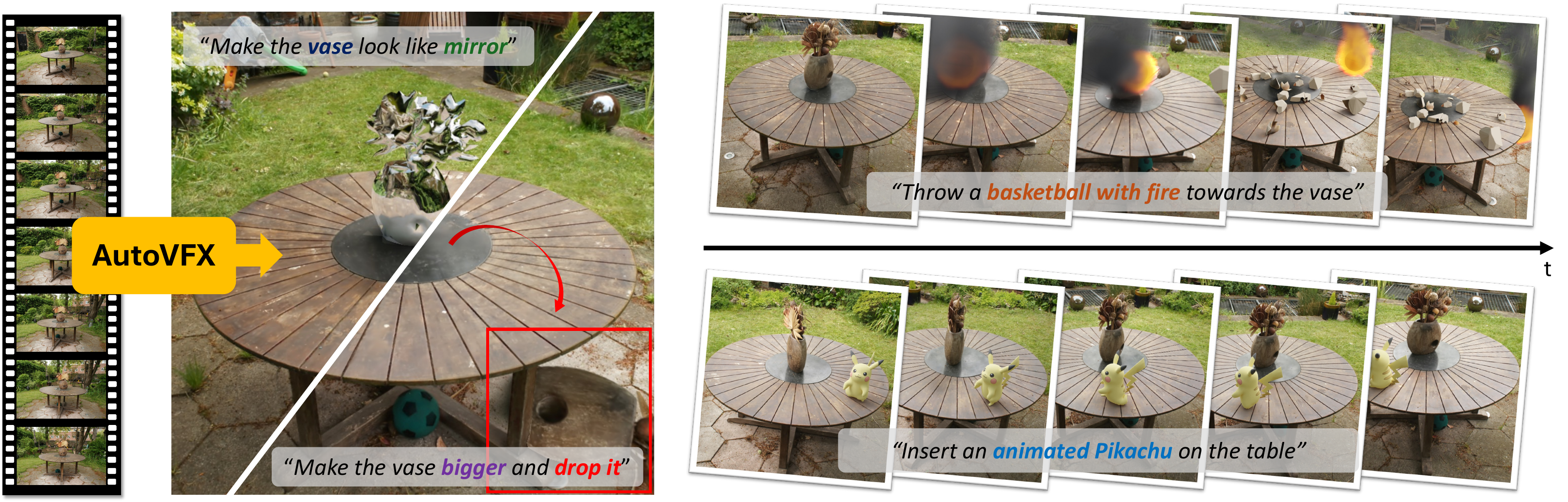}
    \vspace{-5mm}
    \captionof{figure}{{\bf AutoVFX} takes a video and language instructions as input, and automatically generates programs to produce visual effects and render a new video according to the instructions. It can modify appearance and geometry, enable dynamic interactions, apply particle effects, and even insert animated characters, producing results that are photorealistic, physically-plausible, and easily controllable.}
    \label{fig:teaser}
\end{center}

}]

\begin{abstract}
Modern visual effects (VFX) software has made it possible for skilled artists to create imagery of virtually anything. However, the creation process remains laborious, complex, and largely inaccessible to everyday users. In this work, we present \method{}, a framework that automatically creates realistic and dynamic VFX videos from a single video and natural language instructions. By carefully integrating neural scene modeling, LLM-based code generation, and physical simulation, \method{} is able to provide physically-grounded, photorealistic editing effects that can be controlled directly using natural language instructions. We conduct extensive experiments to validate \method{}'s efficacy across a diverse spectrum of videos and instructions. Quantitative and qualitative results suggest that \method{} outperforms all competing methods by a large margin in generative quality, instruction alignment, editing versatility, and physical plausibility.
\end{abstract}

\begin{figure*}[t]
    \centering
    \includegraphics[width=0.9\textwidth]{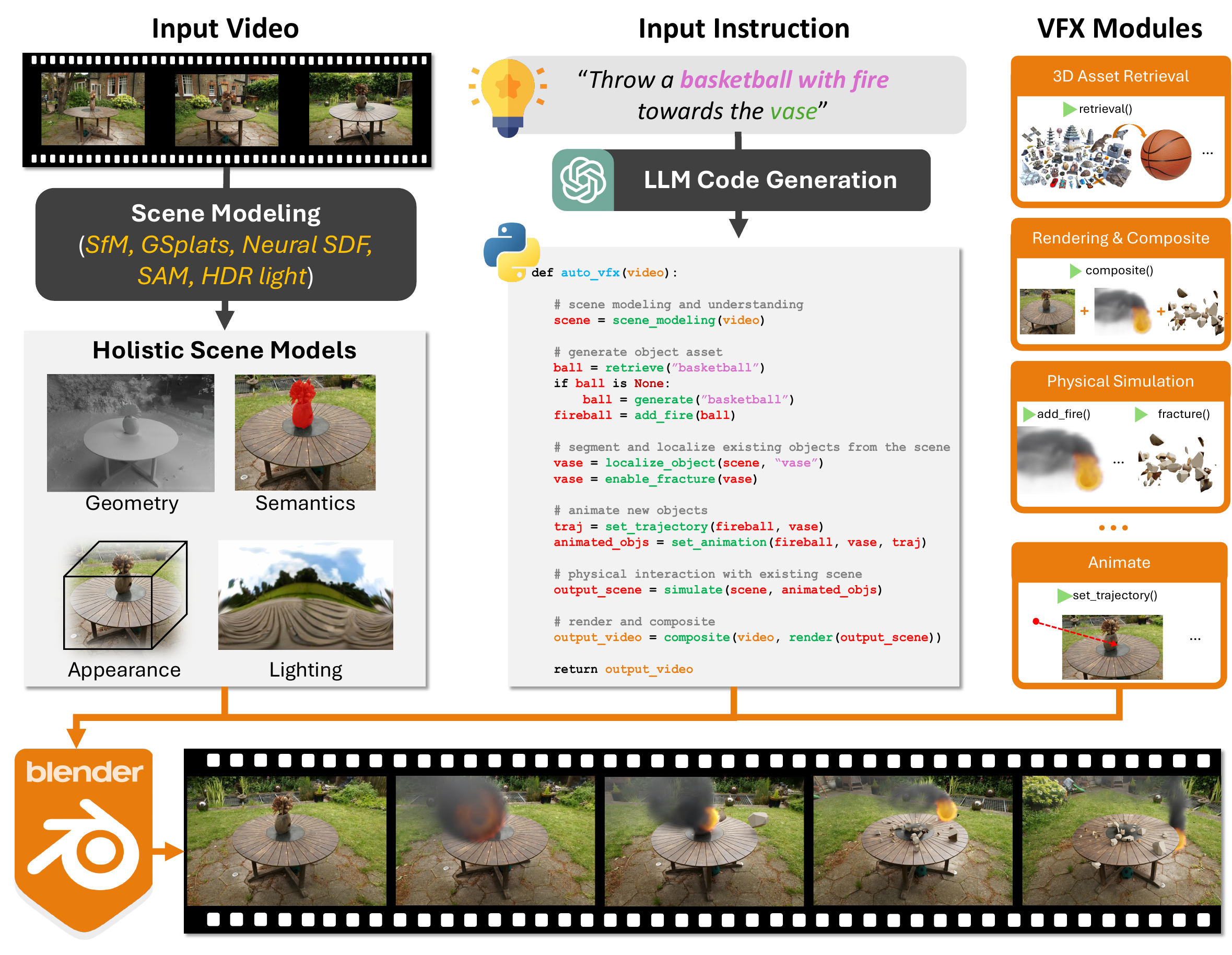}
    \vspace{-3mm}
    \caption{\textbf{AutoVFX framework.} Our instruction-guided video editing framework consists of three main modules: (1) \textbf{3D Scene Modeling} (left), which integrates 3D reconstruction and scene understanding models; (2) \textbf{Program Generation} (middle), where LLMs generate editing programs based on user instructions; and (3) \textbf{VFX Modules} (right), which include predefined functions specialized for various editing tasks. These components are integrated with a physically-based simulation and rendering engine (e.g., Blender) to generate the final video. %
    }
    \label{fig:method}
    \vspace{-5mm}
\end{figure*}

\section{Introduction}

Visual effects (VFX) combine realistic video footage with computer-generated imagery to create novel, photorealistic visuals. Recent advances in graphics, vision, and physical simulation have made it possible to produce VFX that depict virtually anything—even those that are too costly, time-consuming, dangerous, or impossible to capture in real life. As a result, VFX have become essential in modern filmmaking, ads, simulation, AR/VR, \etc. However, the process remains laborious, complex, and expensive, requiring expert skills and professional software~\cite{blender, houdini, christiansen2013adobe, maya}, making it largely inaccessible to everyday users.
 
A promising approach to democratizing VFX is to treat it as a generative video editing problem, where raw video and language prompts are used to generate new videos reflecting the original content and given instructions~\cite{bar2022text2live, wu2023tune, qi2023fatezero, cong2023flatten, tokenflow2023, ceylan2023pix2video, wu2023fairy, ku2024anyv2v, mou2024revideoremakevideomotion, yang2024fresco, ouyang2023codef}. This method leverages advances in generative modeling, learning from large-scale internet data to produce controllable video. Successes have been seen in deepfake videos, fashion, driving, and robotics~\cite{choi2024improving, gao2023magicdrive, sun2024drivescenegen, li2023drivingdiffusion, yang2023learning}. However, this purely data-driven generative editing approach hasn't yet replaced traditional VFX pipelines due to challenges in achieving guaranteed physical plausibility, precise 3D-aware control, and various special effects. %

Another appealing alternative is to build a 3D representation from video input, apply edits like object insertion or texture changes, and then render the final output~\cite{Li2023ClimateNeRF, qiao2023dmrf, instructnerf2023, dong2023vica, mirzaei2023watch, zhuang2024tip, zhuang2023dreameditor, Chen_2024_CVPR, GaussianEditor, chen2024dge, xu2024tiger, fang2024chat, lin2023urbanir, shen2023sim, lin2024iris, liang2024photorealistic}. While this approach aligns well with the VFX pipeline, it is often limited in editing capabilities and still requires manual interaction with cumbersome interfaces, making it difficult for everyday users. Bridging this gap is essential to make 3D scene editing capable of handling most visual effects while remaining accessible to everyone.

In this work, we present \method{}, a framework that automatically creates realistic and dynamic VFX videos from a single video and natural language instructions. At the core of our method is a novel integration of neural scene modeling, LLM-based code generation, and physical simulation. First, we establish a holistic scene model that encodes rich geometry, appearance, and semantics from the input video. This model serves as the foundation for a variety of scene editing, simulation, and rendering capabilities, which we organize into a collection of executable functions. Next, \method{} takes simple language editing instructions and converts them into programs using large language models (LLMs). These programs consist of a sequence of calls to our predefined functions. Finally, the generated code is executed, producing a free-viewpoint video that reflects the instructed changes. Fig.~\ref{fig:method} illustrates the overall framework.

\method{} combines the strengths of generative editing and physical simulation, yet is uniquely set apart from both. Like traditional VFX, \method{} produces videos with physics-grounded, controllable, and photorealistic effects. At the same time, similar to generative editing, we support open-world natural language instructions, allowing anyone to edit a video by simply describing the desired effects. 

We conduct extensive experiments to validate \method{}'s efficacy across a diverse spectrum of videos and instructions. We also perform user studies and qualitative and quantitative comparisons with existing video and scene editing methods. Experimental results suggest \method{} outperforms all competing methods by a large margin 
in generative quality, instruction alignment, editing versatility, and physical plausibility. This demonstrates the effectiveness and convenience of our approach, highlighting its potential as a valuable framework to democratize VFX and pave the way for future integration of even more capabilities to further enhance realism in automatic VFX.

\begin{table*}[!ht]
\renewcommand{\arraystretch}{1.2} %
    \caption{{\bf Comparison of existing and proposed methods for visual editing.} Generative editing models lack physical plausibility and precise controllability. Existing physics-based editing methods have complicated interfaces and are limited in their range of editing capacities. Our method, AutoVFX, enjoys a convenient natural language interface while providing the widest range of capabilities. 
    }
    \vspace{-3mm}
    \centering\setlength{\tabcolsep}{8pt}
    \resizebox{1.0\linewidth}{!}{%
\begin{tabular}{c|cc|cc|ccccccc}
\toprule
\multirow{3}{*}{Method} & \multicolumn{4}{c|}{Input \& Output}                                                                                                                                                                                                                             & \multicolumn{7}{c}{Editing Capacities}                                                                                                                                                                                                                                                     \\ \cline{2-12} 
                        & \multicolumn{1}{c|}{\begin{tabular}[c]{@{}c@{}}Real World\\ Video Editing\end{tabular}} & \begin{tabular}[c]{@{}c@{}}Free-Viewpoint\\ Rendering\end{tabular} & \multicolumn{1}{c|}{\begin{tabular}[c]{@{}c@{}}Editing \\ Interface \\ \end{tabular}} & \begin{tabular}[c]{@{}c@{}}Open-world \\ Query\end{tabular} & \multicolumn{1}{c|}{\begin{tabular}[c]{@{}c@{}}Object \\ Insertion\end{tabular}} & \multicolumn{1}{c|}{\begin{tabular}[c]{@{}c@{}}Object \\ Removal\end{tabular}} & \multicolumn{1}{c|}{\begin{tabular}[c]{@{}c@{}}Object \\ Rearrange\end{tabular}} & \multicolumn{1}{c|}{\begin{tabular}[c]{@{}c@{}}Appearance \\ Change\end{tabular}} & \multicolumn{1}{c|}{\begin{tabular}[c]{@{}c@{}}Animated \\ Objects\end{tabular}} & \multicolumn{1}{c|}{\begin{tabular}[c]{@{}c@{}}Physics\\ Simulation\end{tabular}} & \begin{tabular}[c]{@{}c@{}}Particle \\ Effects\end{tabular} \\ \hline

Visual Programming \cite{gupta2023visual} & \multicolumn{1}{c|}{\greenmark} & \redcross & \multicolumn{1}{c|}{Natural Language} & \greenmark & \multicolumn{1}{c|}{\greenmark} & \multicolumn{1}{c|}{\greenmark} & \multicolumn{1}{c|}{\greenmark} & \multicolumn{1}{c|}{\greenmark} & \multicolumn{1}{c|}{\redcross} & \multicolumn{1}{c|}{\redcross} & \redcross \\

FRESCO \cite{yang2024fresco} & \multicolumn{1}{c|}{\greenmark} & \redcross & \multicolumn{1}{c|}{Natural Language} & \greenmark & \multicolumn{1}{c|}{\redcross} & \multicolumn{1}{c|}{\redcross} & \multicolumn{1}{c|}{\redcross} & \multicolumn{1}{c|}{\greenmark} & \multicolumn{1}{c|}{\redcross} & \multicolumn{1}{c|}{\redcross} & \redcross \\

ClimateNeRF \cite{Li2023ClimateNeRF} & \multicolumn{1}{c|}{\greenmark} & \greenmark & \multicolumn{1}{c|}{Predefined Scripts} & \redcross & \multicolumn{1}{c|}{\redcross} & \multicolumn{1}{c|}{\redcross} & \multicolumn{1}{c|}{\redcross} & \multicolumn{1}{c|}{\greenmark} & \multicolumn{1}{c|}{\redcross} & \multicolumn{1}{c|}{\redcross} & \greenmark \\

Feature Splatting \cite{qiu-2024-featuresplatting} & \multicolumn{1}{c|}{\greenmark} & \greenmark & \multicolumn{1}{c|}{Predefined Scripts} & \greenmark & \multicolumn{1}{c|}{\greenmark} & \multicolumn{1}{c|}{\greenmark} & \multicolumn{1}{c|}{\greenmark} & \multicolumn{1}{c|}{\greenmark} & \multicolumn{1}{c|}{\greenmark} & \multicolumn{1}{c|}{\greenmark} & \redcross \\

GaussianEditor \cite{Chen_2024_CVPR} & \multicolumn{1}{c|}{\greenmark} & \greenmark & \multicolumn{1}{c|}{Graphical} & \greenmark & \multicolumn{1}{c|}{\greenmark} & \multicolumn{1}{c|}{\greenmark} & \multicolumn{1}{c|}{\redcross} & \multicolumn{1}{c|}{\greenmark} & \multicolumn{1}{c|}{\redcross} & \multicolumn{1}{c|}{\redcross} & \redcross \\

Gaussian Grouping \cite{gaussian_grouping} & \multicolumn{1}{c|}{\greenmark} & \greenmark & \multicolumn{1}{c|}{Graphical} & \greenmark & \multicolumn{1}{c|}{\greenmark} & \multicolumn{1}{c|}{\greenmark} & \multicolumn{1}{c|}{\greenmark} & \multicolumn{1}{c|}{\greenmark} & \multicolumn{1}{c|}{\redcross} & \multicolumn{1}{c|}{\redcross} & \redcross \\

PhysGaussian \cite{xie2023physgaussian} & \multicolumn{1}{c|}{\greenmark} & \greenmark & \multicolumn{1}{c|}{Graphical} & \redcross & \multicolumn{1}{c|}{\redcross} & \multicolumn{1}{c|}{\redcross} & \multicolumn{1}{c|}{\redcross} & \multicolumn{1}{c|}{\redcross} & \multicolumn{1}{c|}{\redcross} & \multicolumn{1}{c|}{\greenmark} & \redcross \\

VR-GS \cite{jiang2024vr-gs} & \multicolumn{1}{c|}{\greenmark} & \greenmark & \multicolumn{1}{c|}{Graphical} & \redcross & \multicolumn{1}{c|}{\greenmark} & \multicolumn{1}{c|}{\greenmark} & \multicolumn{1}{c|}{\greenmark} & \multicolumn{1}{c|}{\redcross} & \multicolumn{1}{c|}{\redcross} & \multicolumn{1}{c|}{\greenmark} & \redcross \\

Gaussian Splashing \cite{feng2024splashing} & \multicolumn{1}{c|}{\greenmark} & \greenmark & \multicolumn{1}{c|}{Graphical} & \redcross & \multicolumn{1}{c|}{\greenmark} & \multicolumn{1}{c|}{\greenmark} & \multicolumn{1}{c|}{\greenmark} & \multicolumn{1}{c|}{\redcross} & \multicolumn{1}{c|}{\redcross} & \multicolumn{1}{c|}{\greenmark} & \greenmark \\

DMRF \cite{qiao2023dmrf} & \multicolumn{1}{c|}{\greenmark} & \greenmark & \multicolumn{1}{c|}{Graphical} & \redcross & \multicolumn{1}{c|}{\greenmark} & \multicolumn{1}{c|}{\redcross} & \multicolumn{1}{c|}{\redcross} & \multicolumn{1}{c|}{\redcross} & \multicolumn{1}{c|}{\greenmark} & \multicolumn{1}{c|}{\greenmark} & \greenmark \\

Instruct-N2N \cite{instructnerf2023} & \multicolumn{1}{c|}{\greenmark} & \greenmark & \multicolumn{1}{c|}{Natural Language} & \greenmark & \multicolumn{1}{c|}{\redcross} & \multicolumn{1}{c|}{\redcross} & \multicolumn{1}{c|}{\redcross} & \multicolumn{1}{c|}{\greenmark} & \multicolumn{1}{c|}{\redcross} & \multicolumn{1}{c|}{\redcross} & \redcross \\

DGE \cite{chen2024dge} & \multicolumn{1}{c|}{\greenmark} & \greenmark & \multicolumn{1}{c|}{Natural Language} & \greenmark & \multicolumn{1}{c|}{\redcross} & \multicolumn{1}{c|}{\redcross} & \multicolumn{1}{c|}{\redcross} & \multicolumn{1}{c|}{\greenmark} & \multicolumn{1}{c|}{\redcross} & \multicolumn{1}{c|}{\redcross} & \redcross \\

Chat-Edit-3D \cite{fang2024chat} & \multicolumn{1}{c|}{\greenmark} & \greenmark & \multicolumn{1}{c|}{Natural Language} & \greenmark & \multicolumn{1}{c|}{\greenmark} & \multicolumn{1}{c|}{\greenmark} & \multicolumn{1}{c|}{\greenmark} & \multicolumn{1}{c|}{\greenmark} & \multicolumn{1}{c|}{\redcross} & \multicolumn{1}{c|}{\redcross} & \redcross \\

ChatSim \cite{wei2024editable} & \multicolumn{1}{c|}{\greenmark} & \greenmark & \multicolumn{1}{c|}{Natural Language} & \redcross & \multicolumn{1}{c|}{\greenmark} & \multicolumn{1}{c|}{\greenmark} & \multicolumn{1}{c|}{\greenmark} & \multicolumn{1}{c|}{\greenmark} & \multicolumn{1}{c|}{\greenmark} & \multicolumn{1}{c|}{\redcross} & \redcross \\

\textbf{\method (Ours)} & \multicolumn{1}{c|}{\greenmark} & \greenmark & \multicolumn{1}{c|}{Natural Language} & \greenmark & \multicolumn{1}{c|}{\greenmark} & \multicolumn{1}{c|}{\greenmark} & \multicolumn{1}{c|}{\greenmark} & \multicolumn{1}{c|}{\greenmark} & \multicolumn{1}{c|}{\greenmark} & \multicolumn{1}{c|}{\greenmark} & \greenmark \\ \bottomrule                                                    

\end{tabular}
}
\vspace{-5mm}
\label{tab:related_works}
\end{table*}

\section{Related Work}

Our framework is closely related to several areas, including physical simulation on NeRFs, instruction-guided visual editing, and LLMs for code generation, integrating aspects of all three. Next, we will discuss these areas and highlight and contrast notable works in Tab.~\ref{tab:related_works}.

\paragraph{Physical Simulation on NeRFs and 3D Gaussians}

Integrating physics simulation into NeRFs and 3D Gaussians enables immersive and convincing dynamic effects within captured scenes. Several lines of work have explored various physical interactions, including rigid body object interaction~\cite{xia2024video2game, wei2024editable}, particle physics effects such as flooding and fog~\cite{feng2024splashing, Li2023ClimateNeRF}, elastic deformable objects~\cite{zhang2024physdreamer}, and plastic objects~\cite{jiang2024vr-gs, xie2023physgaussian, qiao2022neuphysics}. The key idea is to enable captured scenes to faithfully interact with new events or entities through physical simulation. However, this is challenging for vanilla neural implicit models, as conventional simulation often requires high-fidelity surface geometry, which is not explicit in these models. Therefore, various approaches seek to extract meshes from NeRFs~\cite{neumesh, xu2022deforming, peng2022cagenerf, yuan2022nerf, qiao2022neuphysics, jiang2024vr-gs} to facilitate simulation, while others adapt implicit or particle-based simulation so that it can be directly applied to implicit models or Gaussians~\cite{li2023pacnerf, feng2023pienerf, xie2023physgaussian, feng2024splashing}. AutoVFX explores a hybrid representation where meshes are used for physical interaction and Gaussians are stacked on the mesh surface for rendering, combining the best of both worlds. Another challenge is that some physical interactions require an understanding of physical properties. Various approaches address this through inverse physics~\cite{li2023pacnerf, zhang2024physdreamer}, common sense knowledge in large foundation models~\cite{zhai2024physical, liu2025physgen, fu2024blink}, or generative models~\cite{zhang2024physdreamer}. Most works on physical simulation, however, are driven by domain-specific scripts rather than natural language instructions, often restricting them to specific physical effects and limiting their user base. AutoVFX seeks to bridge this gap by using LLMs to convert language instructions into simulation programs and supporting numerous dynamical effects through off-the-shelf simulators.

\paragraph{Instruction-based Visual Editing}

Recent advancements in visual-language models have made visual editing more accessible by allowing users to edit a wide range of content, such as images, videos, and 3D scenes, using language instructions~\cite{instructnerf2023, brooks2022instructpix2pix, yang2024fresco, ku2024anyv2v, fu2024blink} instead of relying on GUI interactions or script programs~\cite{Chen_2024_CVPR, jiang2024vr-gs, qiu-2024-featuresplatting, gaussian_grouping, Li2023ClimateNeRF}. Text- and image-conditioned generative models, particularly diffusion-based approaches~\cite{ho2020denoising, rombach2021highresolution}, have been explored for text-guided image~\cite{svd, rombach2021highresolution, meng2022sdedit, zhang2023adding, brooks2022instructpix2pix} and video~\cite{yang2024fresco, bar2022text2live, wu2023tune, qi2023fatezero, cong2023flatten, tokenflow2023, ceylan2023pix2video, wu2023fairy, ku2024anyv2v, mou2024revideoremakevideomotion, ouyang2023codef} editing. Language-embedded NeRFs extend this generative editing capability to 3D scenes~\cite{instructnerf2023, GaussianEditor, Chen_2024_CVPR, chen2024dge, mirzaei2023watch, dong2023vica, zhuang2023dreameditor, zhuang2024tip, xu2024tiger, song2023efficient}. However, mapping text instructions to desired edits in videos and scenes purely through diffusion models can be challenging, particularly when tasks involve complex steps, dynamic interactions, or physics, which can affect instruction alignment, physical plausibility, or realism. To address this, some methods use large language models (LLMs) to break down tasks into subtasks~\cite{wei2024editable, fang2024chat, GaussianEditor} or generate executable programs based on instructions~\cite{gupta2023visual, Lv_2024_CVPR}. AutoVFX belongs to this latter category but demonstrates significantly richer capabilities, such as dynamic visual effects, animated objects and physical interaction, compared to these methods.

\paragraph{LLMs for Code Generation}
The powerful capabilities of large language models (LLMs) have revolutionized code generation based on natural language descriptions. By providing in-context examples, LLMs can generate code snippets in specific formats or syntaxes. Studies such as \cite{drori2022neural, chen2021evaluating, austin2021program} have explored the effectiveness of LLMs in solving math and code problems. LLM-based code generation has recently been investigated in vision and robotics. Many works adopt LLMs for decision-making in embodied AI, including tasks like manipulation~\cite{singh2023progprompt, huang2023voxposer, liang2023code, yang2023octopus} and navigation~\cite{ma2024lampilot}. Recently, LLM-based programs for visual content creation have gained attention, with notable progress in 2D understanding and editing~\cite{gupta2023visual}, driving simulation~\cite{fang2024chat}, video generation~\cite{Lv_2024_CVPR}, and procedural 3D scene generation~\cite{hu2024scenecraft, raistrick2023infinite, raistrick2024infinigen, zhou2024scenex, gao2024graphdreamer}. In our method, we harness GPT-4 to interpret natural language descriptions into executable programs for creating diverse visual effects for generic real-world videos.

\begin{figure}
    \centering
    \includegraphics[width=.95\linewidth]{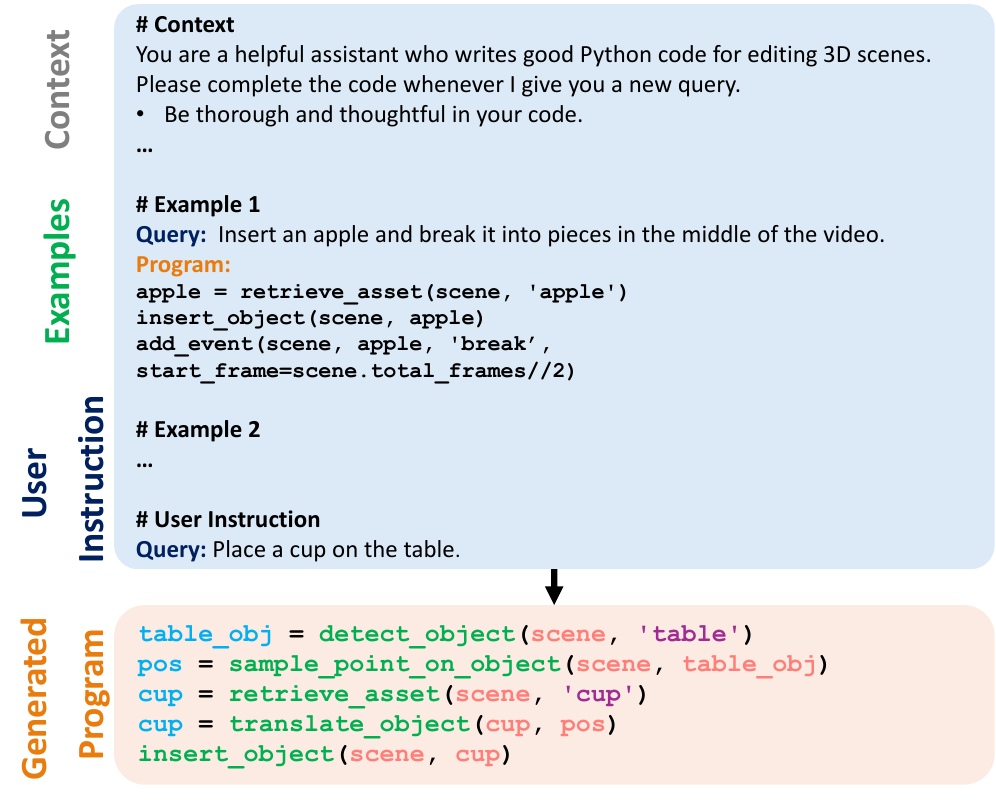}
    \vspace{-1mm}
    \caption{\textbf{Program generation.} The LLM generates the editing program through in-context learning. With provided context and examples, it learns to call VFX modules and, given unseen user instructions (blue block), generates the program (orange block).
    }
    \label{fig:program_generation}
    \vspace{-5mm}
\end{figure}

\section{Text-Driven VFX Creator}
\method{} takes as input a video and a natural language editing prompt, and outputs an edited free-viewpoint video.
The core idea is to uniquely combine the code generation capabilities of LLMs with 3D scene modeling and physics-based simulation techniques. Fig.~\ref{fig:method} depicts the overall framework. First, we harness various 3D vision methods to estimate key scene properties from the input video (Sec.~\ref{sec:modeling}). This lays the foundation for a variety of scene editing, simulation, and rendering capabilities, which we organize into a collection of executable modules (Sec.~\ref{sec:editing}, Sec.~\ref{sec:rendering}). An LLM is used to convert the natural language editing instructions into a program calling these functions (Sec.~\ref{sec:llm_agent}). Finally, the generated program is executed, producing a free-viewpoint video that reflects the instructed changes.

\subsection{3D Scene Modeling}
\label{sec:modeling}
Photo-realistic, physics-based VFX creation requires modeling several key properties of the captured scene, namely geometry, appearance, semantics, and lighting. We employ a variety of recent scene understanding models to estimate these properties, including separate models for geometry and appearance in order to achieve both high simulation fidelity and photorealistic rendering.

\paragraph{Geometry} Modeling the 3D geometry of the scene is essential for any sort of object insertion, removal, or simulation. We first run COLMAP~\cite{schonberger2016structure} to infer the camera poses of each frame.
We then capture the geometry in the form of a triangle mesh produced by BakedSDF~\cite{yariv2023bakedsdf}, a multi-view reconstruction method that optimizes a hybrid implicit neural representation that encodes the signed distance field of the scene, and then bakes the representation onto a triangulated mesh. It achieves a desirable balance between surface accuracy, completeness, and efficiency.
We choose to use a mesh representation because it can be directly loaded into a standard VFX pipeline, rendered efficiently, and further enables accurate physical simulation. Moreover, it serves as the geometry proxy for object instance extraction.

\paragraph{Appearance} We capture the appearance of the scene in two ways. First, we use SuGaR~\cite{guedon2023sugar}, a Gaussian Splatting~\cite{kerbl20233d} based novel view synthesis method, to enable freeview rendering. Although SuGaR can provide realistics renderings, it cannot be directly incorporated into physical-based rendering, making it unsuitable for modeling reflective effects on inserted objects or material editing. 
Thus, we also represent the scene by texturing the BakedSDF mesh, which has lower visual fidelity but can be integrated with physically-based rendering. This textured mesh is used for shadow mapping and encoding multi-bounce effects.

\paragraph{Semantics} In many cases, a user would like to perform an edit localized to a specific semantic region of the scene, for example ``make the car on fire''. To enable such edits, we use Grounding SAM~\cite{liu2023grounding} to perform open-vocabulary instance segmentation and DEVA~\cite{cheng2023tracking} to associate the instances across frames. To lift video segmentation to 3D, we first un-project each pixel from the 2D segmentation mask into the 3D scene geometry. A voting mechanism is used to determine the visibility of mesh vertices across multiple camera views. By setting a threshold for visibility, we select mesh faces that meet or exceed this threshold. Next, we find 3D Gaussians that are closest to these selected mesh faces and render them to produce an alpha image. We then calculate the average mean Intersection over Union (mIoU) between the rendered alpha image and the original segmentation masks. Finally, we select the mesh faces and 3D Gaussians with the highest mIoU, representing the most accurate 3D segmentation.

\paragraph{Lighting}
Accurate lighting estimation ensures that all elements within the scene are coherently illuminated. We estimate the environmental lighting of a scene in two ways. For fully captured indoor scenes, such as those in ScanNet++~\cite{yeshwanth2023scannet++}, we unproject the over-saturated image pixels into space and use majority voting to determine the estimated emitter meshes. These meshes with emissions lights are subsequently imported into the renderer to serve as light sources. For partially captured indoor scenes and outdoor scenes, such as those in MipNeRF360~\cite{barron2022mip}, we use DiffusionLight~\cite{phongthawee2023diffusionlight} to inpaint chrome balls in the center of initial frame at multiple exposure levels. An high dynamic range (HDR) map is then generated from these inpainted frames and imported into the renderer as an environmental light.

\begin{figure*}[t]
    \centering
    \setlength\tabcolsep{0.1em}
    \resizebox{1.0\textwidth}{!}{%
        \begin{tabular}{@{}lcccccc@{}}

        & \multicolumn{3}{c}{\prompt{Simulate books falling from the sofa.}} & \multicolumn{3}{c}{\prompt{Insert an animated dragon moving around the floor.}} \\

       & \includegraphics[width=0.16\textwidth]{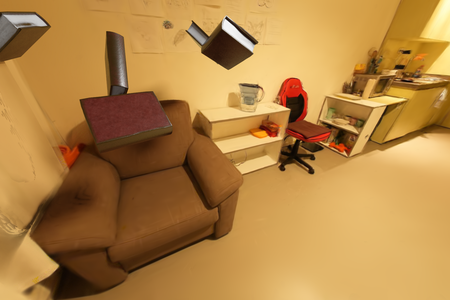} & \includegraphics[width=0.16\textwidth]{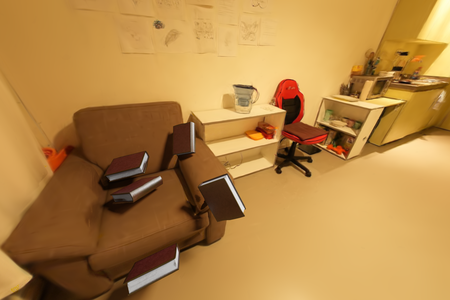} & \includegraphics[width=0.16\textwidth]{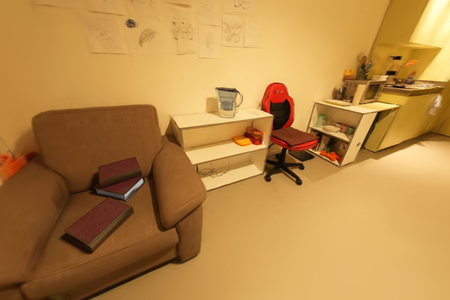} & \includegraphics[width=0.16\textwidth]{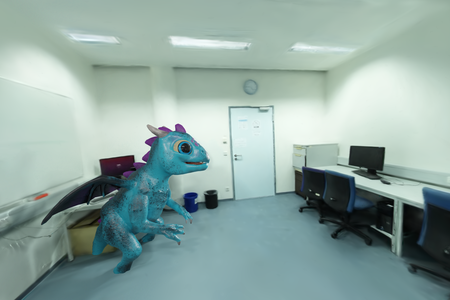} & \includegraphics[width=0.16\textwidth]{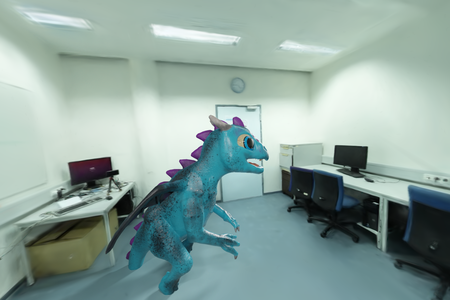} & \includegraphics[width=0.16\textwidth]{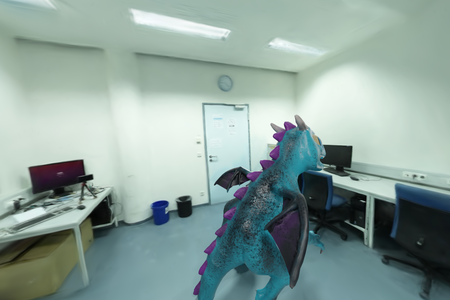} \\
       
       & \multicolumn{3}{c}{\prompt{Insert computer on the table emitting smoke.}} & \multicolumn{3}{c}{\prompt{Setup camp fire on the floor.}} \\

       & \includegraphics[width=0.16\textwidth]{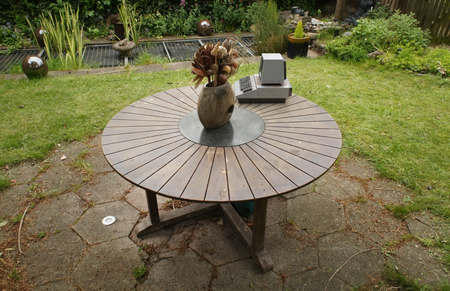} & \includegraphics[width=0.16\textwidth]{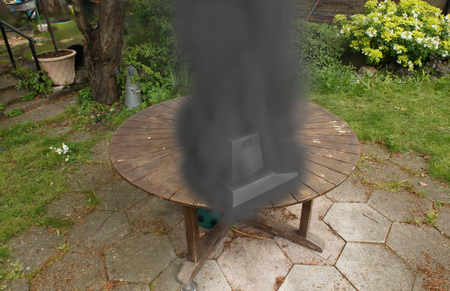} & \includegraphics[width=0.16\textwidth]{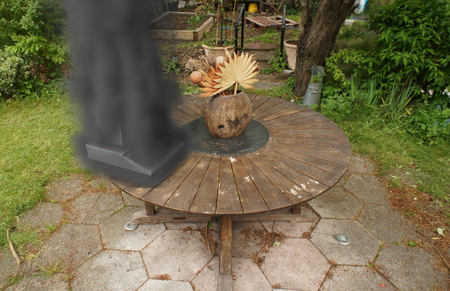} & \includegraphics[width=0.16\textwidth]{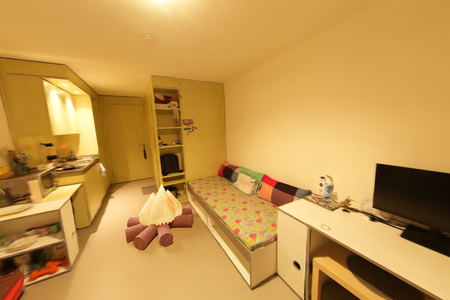} & \includegraphics[width=0.16\textwidth]{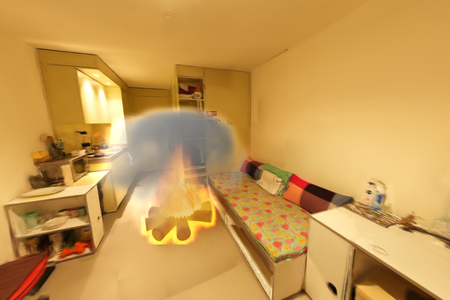} & \includegraphics[width=0.16\textwidth]{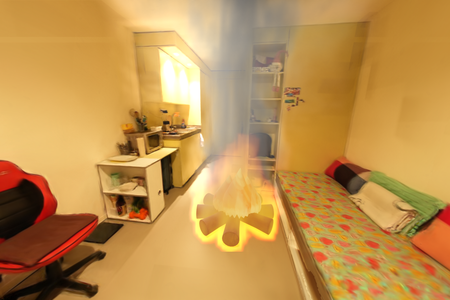} \\

        & \multicolumn{3}{c}{\prompt{Insert a mirrored Pikachu and a metallic Pikachu on the table.}} & \multicolumn{3}{c}{\prompt{Generate a smiling sunflower on the sink.}} \\

       & \includegraphics[width=0.16\textwidth]{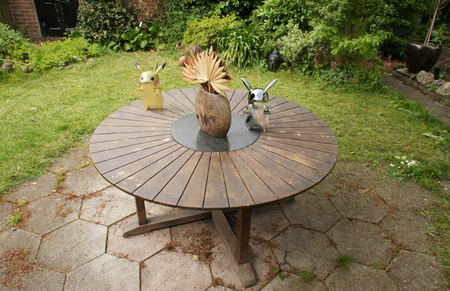} & \includegraphics[width=0.16\textwidth]{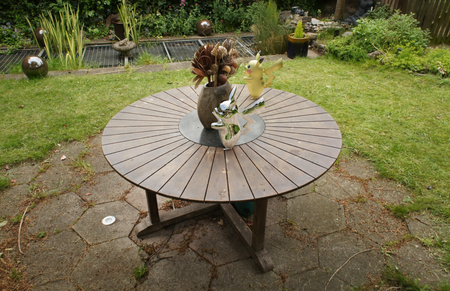} & \includegraphics[width=0.16\textwidth]{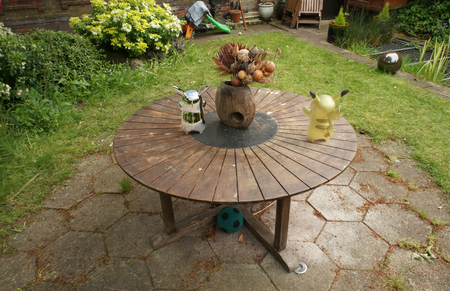} & \includegraphics[width=0.16\textwidth]{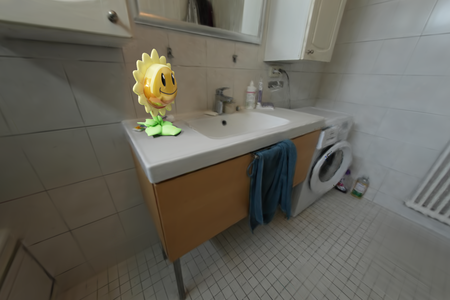} & \includegraphics[width=0.16\textwidth]{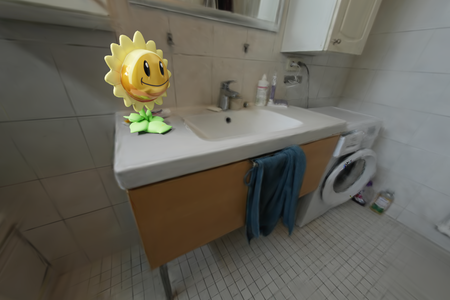} & \includegraphics[width=0.16\textwidth]{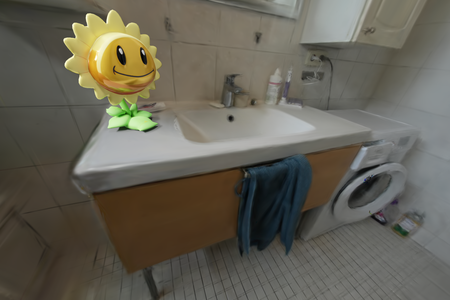} \\

       & \multicolumn{3}{c}{\prompt{Drop four barrels onto the floor with texture modified.}} & \multicolumn{3}{c}{\prompt{Break the sculpture.}} \\

       & \includegraphics[width=0.16\textwidth]{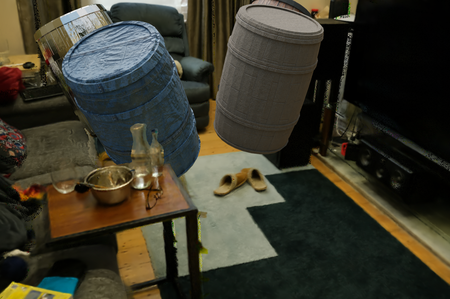} & \includegraphics[width=0.16\textwidth]{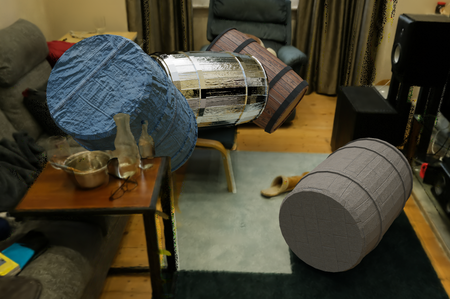} & \includegraphics[width=0.16\textwidth]{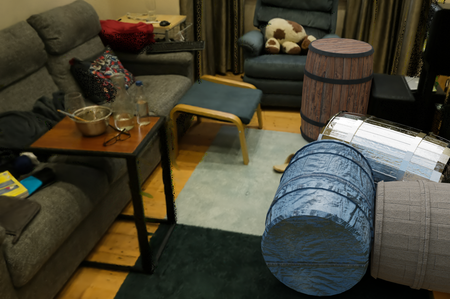} & \includegraphics[width=0.16\textwidth]{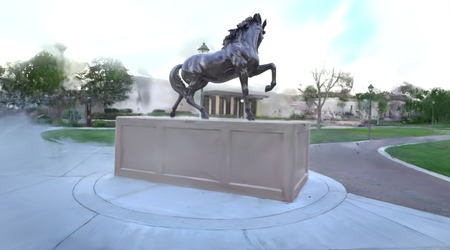} & \includegraphics[width=0.16\textwidth]{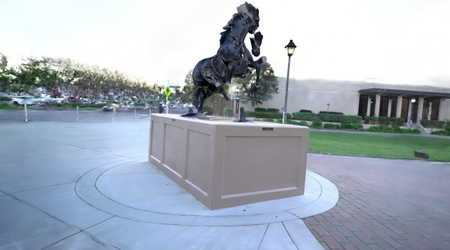} & \includegraphics[width=0.16\textwidth]{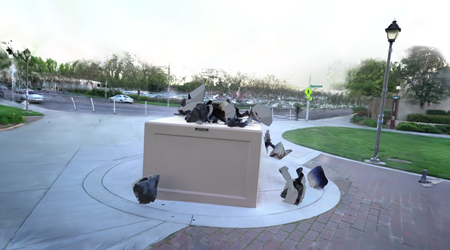} \\
       
    \end{tabular}    
    }
    \caption{{\bf Dynamic VFX video editing using \method.} Our approach enables physical interaction, articulated animation, particle effects, insertion of generated 3D assets, material editing, and geometry fracturing.}
    \vspace{-5mm}
    \label{fig:qual_extra}
\end{figure*}

\subsection{Scene Editing and Simulation}
\label{sec:editing}
The multimodal 3D scene modeling described above paves the way for a wide assortment of editing, simulation, and rendering operations to be performed. We design a suite of intuitive modules that can be seamlessly composed together to provide a rich set of VFX capabilities. We note that this modular framework also allows new capabilities to be easily added via registering new modules. We describe the specific techniques used within each module below.

\paragraph{3D Asset Creation}
To enable diverse object insertions, we use the Objaverse 1.0 dataset~\cite{deitke2023objaverse} and a high-quality subset from Richdreamer~\cite{qiu2023richdreamer} with 280k annotated 3D assets. We adopt a twofold approach for 3D asset creation. 
We first rank 3D assets using Sentence-BERT~\cite{reimers2019sentence} to match query text and identify the top K candidates, then refine the selection with CLIP~\cite{radford2021learning} based on multi-view renderings to select best aligned asset. For text queries beyond existing descriptions, we use Meshy AI to generate high-quality 3D assets with PBR materials, expanding the range of insertable objects.

\paragraph{Insertion}
To achieve realistic object insertion, two critical properties must be accurately determined: position and scale. To ensure plausible positioning of objects, we sample the centers of triangles from the supporting mesh that are sufficiently flat to provide accurate support for the objects. For scaling, we utilize GPT-4V~\cite{achiam2023gpt} models to estimate the real-world dimensions of 3D assets. Detailed prompts for scale estimation are illustrated in the supplementary material.

\paragraph{Removal}
To effectively remove a specific instance from a scene, we begin by extracting the target objects using semantic modules. We remove these Gaussian points along with their associated mesh faces. For geometric restoration, we employ a planar mesh to cover the exposed area on the bottom. For appearance restoration, we first use LaMa~\cite{suvorov2022resolution} to inpaint the missing regions across all video frames. Then, we fine-tune the current 3D Gaussian Splatting model on the inpainted frames to ensure a 3D-consistent recovery.

\paragraph{Material Editing}
Accurate material editing ensures 3D assets responding correctly to lighting, shading, and environmental conditions, helping them blend seamlessly with live-action footage for convincing visual effects. We provide multiple options for material editing, all of which result in modifications to the material nodes of a 3D asset in the renderer. Users can adjust parameters such as metallic, specular, and roughness values of an asset. Users can also modify the overall color of an asset by altering the color intensity in the texture image of an 3D asset. We also support queries by material name, enabling a search across a material database sourced from PolyHaven. The queried material can then be imported and applied to all relevant material nodes.

\paragraph{Physical Simulation}
Because our scene model is directly compatible with Blender, we can leverage its powerful simulation capabilities by simply calling functions from its simulation library. We use these functions to enable simulations of rigid-body physics and particle effects. Rigid-body physics in Blender is based on the Bullet physics engine~\cite{coumans2015bullet}. To achieve both accurate and realistic object interactions, we pre-compute the center of mass and convex hull for collision checking of any interactive objects. Particle effects, such as smoke and fire, rely upon mantaflow~\cite{mantaflow}. We modify the default simulation settings to ensure convincing effects. Further details are provided in the supplementary.

\subsection{Scene Rendering and Video Compositing}
\label{sec:rendering}
We use a careful rendering and compositing scheme in order to produce a photorealistic video result. First, we render inserted objects while keeping the background mesh invisible to first-bounces during raytracing but visible to higher-order bounces. This allows the rendered objects to be affected by lighting from the background. Next, we set the background mesh to be visible and render twice: with and without the inserted objects. We use the ratio of the pixel values between these two as an approximation of the objects' effects on the background surfaces. This ratio is multiplied to either a SuGaR rendering or the original video frames, depending on if novel views are desired. Finally, we alpha-blend the inserted objects into the video, using depth maps of the background mesh for occlusion reasoning. If the objects contain fire (emissive transparent elements), we use premultiplied alpha-blending; otherwise, we use straight alpha-blending.

\subsection{LLM Integration}
\label{sec:llm_agent}
AutoVFX aims to enable the creation of VFX directly from natural language instructions, providing a user-friendly interface accessible to anyone. Towards this goal, we integrate our editing modules into an API within an LLM-agent framework, drawing inspiration from recent works such as Code-as-Policies~\cite{liang2023code} and Visual Programming~\cite{gupta2023visual}. Leveraging GPT-4~\cite{achiam2023gpt}, we prompt the model with in-context examples that pair editing instructions with corresponding programs composed of our predefined editing functions. The LLM then generates a program that is directly executed to perform the specified scene edits.

\paragraph{Modular Function Encapsulation}
Our method encapsulates predefined editing modules into callable and executable functions, which can be combined to form comprehensive programs. Scene objects are represented as Python dictionaries, facilitating straightforward and interpretable edits. Each function's parameters are fully transparent, allowing users with varying levels of programming expertise to manipulate the process. A full list of the predefined modules is available in the supplementary material.

\paragraph{LLM-driven Program Composition and Execution}
To ensure GPT-4 effectively composes these functions into executable programs, we design prompts that guide the model. These prompts include examples demonstrating the combination of predefined functions into Python-like scripts that represent the desired scene edits. Once generated, the program is executed within a Python interpreter, triggering the associated simulations or rendering processes to produce the final visual effect. As illustrated in Fig.~\ref{fig:program_generation}, this approach simplifies the creation of complex visual effects, making it accessible to a broader audience through natural language instructions. The system's modular design also provides flexibility and scalability, allowing users to customize and extend functionality as needed.

\begin{table*}[!ht]
    \caption{Quantitative comparison with other methods. We employ automatic metrics and human evaluation to evaluate the performance. AutoVFX consistently outperforms baseline methods across various metrics. %
    }
    
    \centering\setlength{\tabcolsep}{10pt}
    \resizebox{1.0\linewidth}{!}{%
        \begin{tabular}{c|ccc|cccc|cc}
            \toprule
            \multirow{3}{*}{Method} & \multicolumn{3}{c|}{\bf Semantic Consistency Measures} & \multicolumn{4}{c|}{\bf Multimodal LLM Quality Evaluation} & \multicolumn{2}{c}{\bf User Study} \\ \cline{2-10} 
            & \multicolumn{1}{c}{\begin{tabular}[c]{@{}c@{}}Object\\ Detection\end{tabular}} & \multicolumn{1}{c}{\begin{tabular}[c]{@{}c@{}}CLIP\\ Similarity\end{tabular}} & \multicolumn{1}{c|}{\begin{tabular}[c]{@{}c@{}}CLIP Directional\\ Similarity\end{tabular}} & \multicolumn{1}{c}{\begin{tabular}[c]{@{}c@{}}Photo-\\ realism\end{tabular}} & \multicolumn{1}{c}{\begin{tabular}[c]{@{}c@{}}Text\\ Alignment\end{tabular}} & \multicolumn{1}{c}{\begin{tabular}[c]{@{}c@{}}Structure\\ Preservation\end{tabular}} & \multicolumn{1}{c|}{\begin{tabular}[c]{@{}c@{}}Overall\\ Quality\end{tabular}} & \multicolumn{1}{c}{\begin{tabular}[c]{@{}c@{}}Text\\ Alignment\end{tabular}} & \multicolumn{1}{c}{\begin{tabular}[c]{@{}c@{}}Video\\ Quality\end{tabular}} \\ \hline
            Instruct-N2N \cite{instructnerf2023} & 0.343 & 0.209 & 0.019 & 0.402 & 0.329 & 0.440 & 0.043 & 0.07 & 0.04 \\
            DGE \cite{chen2024dge} & 0.347 & 0.195 & 0.278 & 0.562 & 0.312 & 0.619 & 0.106 & 0.06 & 0.03 \\
            FRESCO \cite{yang2024fresco} & 0.373 & \textbf{0.214} & 0.009 & 0.622 & 0.427 & 0.632 & 0.204 & 0.04 & 0.02 \\
            AutoVFX (Ours) & \textbf{0.537} & 0.206 & \textbf{0.419} & \textbf{0.735} & \textbf{0.791} & \textbf{0.749} & \textbf{0.647} & \textbf{0.83} & \textbf{0.90} \\
            \hline
        \end{tabular}
    }
    \vspace{-5mm}
    \label{tab:quant_all}
\end{table*}

\section{Experiments}

We evaluate AutoVFX across a diverse set of scenes and editing prompts and provide both qualitative and quantitative comparisons with other related methods.

\subsection{Experimental Details}

\paragraph{Dataset \& Preprocessing}
We adopt scenes from real-world datasets such as Mip-NeRF360~\cite{barron2022mip}, Tanks \& Temples~\cite{knapitsch2017tanks}, ScanNet++~\cite{yeshwanth2023scannet++}, and Waymo dataset~\cite{sun2020scalability} to demonstrate our editing capabilities across diverse scenarios. %
We use COLMAP~\cite{schoenberger2016sfm} to extract camera poses and sparse point clouds from images for GSplat initialization. 

\paragraph{Baselines}
We compare our method with three text-based visual editing methods: Instruct-N2N~\cite{instructnerf2023}, DGE~\cite{chen2024dge} and FRESCO~\cite{yang2024fresco}. Instruct-N2N and DGE both perform edits on 3D scene representations based on text descriptions, with the former utilizing NeRF and the latter relying on 3D Gaussians. FRESCO, on the other hand, translates an input video to align with a target text prompt. For our experiments, we set the guidance scale to 12.5 for DGE to achieve more noticeable edits. Apart from this adjustment, all experimental setups adhere to the default settings of each method to ensure a fair comparison. The qualitative results for Instruct-N2N and DGE are obtained by rendering from the edited 3D representations.

\paragraph{Implementation Details}
To render objects that are affected by rigid-body physics, we store the rigid transformations at each timestep and apply them to the 3D Gaussians during rendering. For animating objects based on keypoints, we use Bézier interpolation to produce a smooth trajectory. Additional implementation details regarding the various VFX modules can be found in the supplementary.

\subsection{Qualitative evaluation}

\paragraph{Qualitative comparison}
\begin{figure*}[t]
    \centering
    \setlength\tabcolsep{0.1em}
    \resizebox{1.0\textwidth}{!}{%
        \begin{tabular}{@{}lccccc@{}}

        & \multicolumn{5}{c}{\prompt{Make vase with flowers to be like a mirror.}} \\
        \raisebox{2.5\normalbaselineskip}{\rotatebox[origin=c]{90}{{Garden}}} & \includegraphics[width=0.195\textwidth]{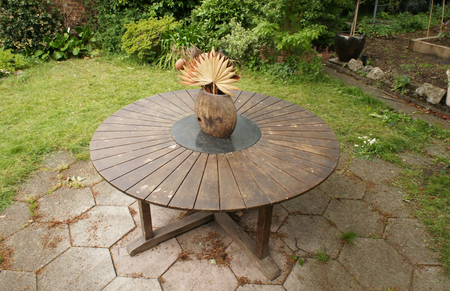} & \includegraphics[width=0.195\textwidth]{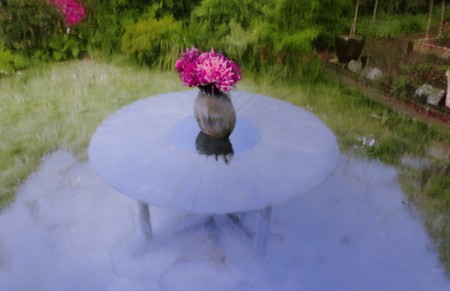} & \includegraphics[width=0.195\textwidth]{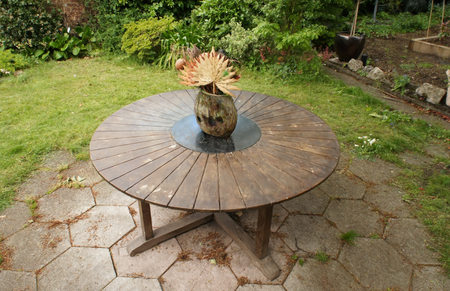} & \includegraphics[width=0.195\textwidth]{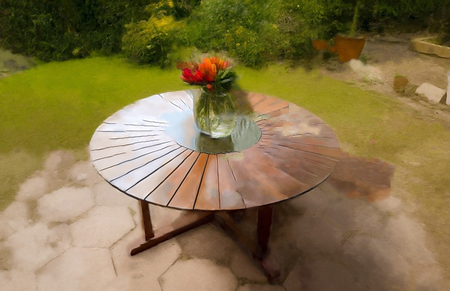} & \includegraphics[width=0.195\textwidth]{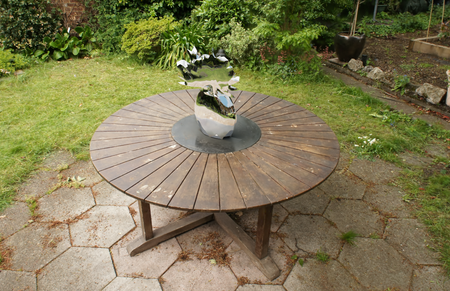} \\

        & \multicolumn{5}{c}{\prompt{Add a cupcake on the metal plate.}} \\
       \raisebox{2.5\normalbaselineskip}[0pt][0pt]{\rotatebox[origin=c]{90}{{Counter}}} & \includegraphics[width=0.195\textwidth]{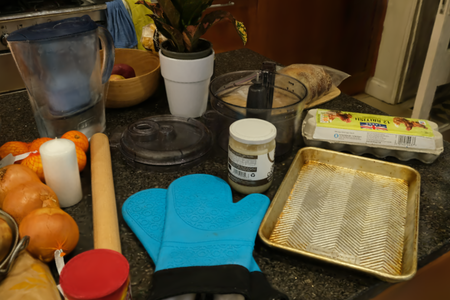} & \includegraphics[width=0.195\textwidth]{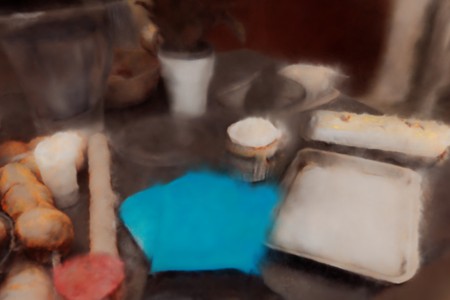} & \includegraphics[width=0.195\textwidth]{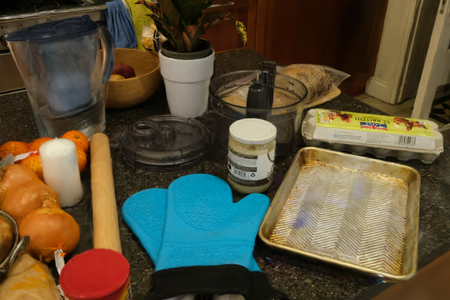} & \includegraphics[width=0.195\textwidth]{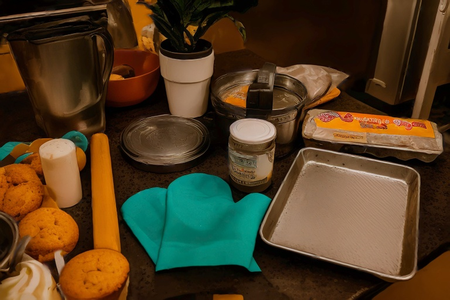} & \includegraphics[width=0.195\textwidth]{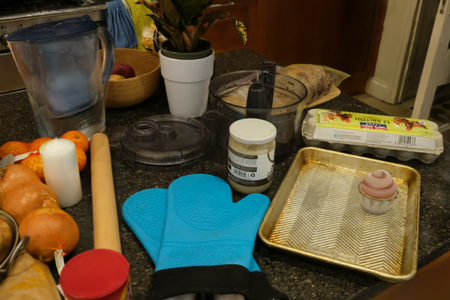} \\
        
        & \multicolumn{5}{c}{\prompt{Replace bulldozer with birthday cake at the same position.}} \\
       \raisebox{2.5\normalbaselineskip}[0pt][0pt]{\rotatebox[origin=c]{90}{{Kitchen}}} & \includegraphics[width=0.195\textwidth]{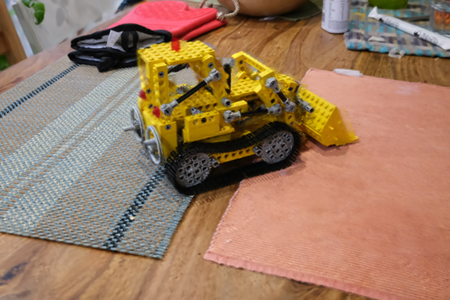} & \includegraphics[width=0.195\textwidth]{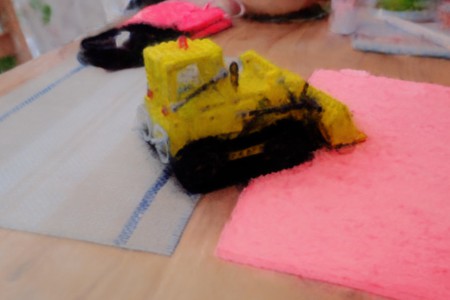} & \includegraphics[width=0.195\textwidth]{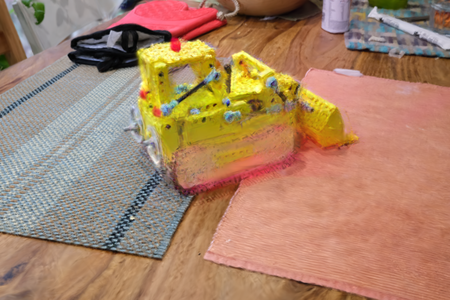} & \includegraphics[width=0.195\textwidth]{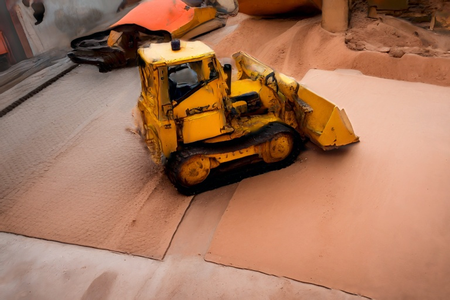} & \includegraphics[width=0.195\textwidth]{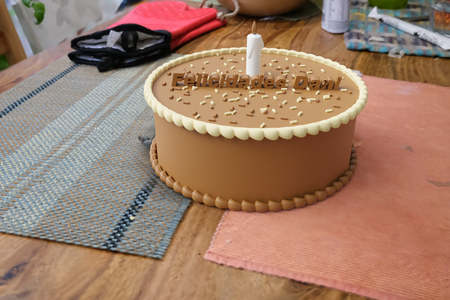} \\

        & \multicolumn{5}{c}{\prompt{Duplicate the bulldozer twice: move one to (0, 0.5, 0) and make it metallic;}} \\
        & \multicolumn{5}{c}{\prompt{move the other to (0.5, 0, 0) and make it mirror-like.}} \\
       \raisebox{2.5\normalbaselineskip}[0pt][0pt]{\rotatebox[origin=c]{90}{{Kitchen}}} & \includegraphics[width=0.195\textwidth]{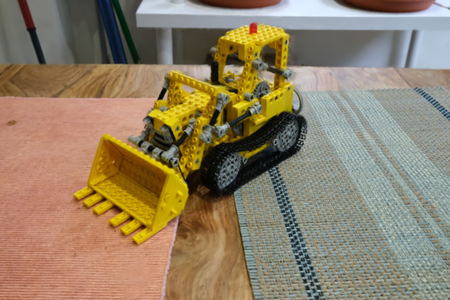} & \includegraphics[width=0.195\textwidth]{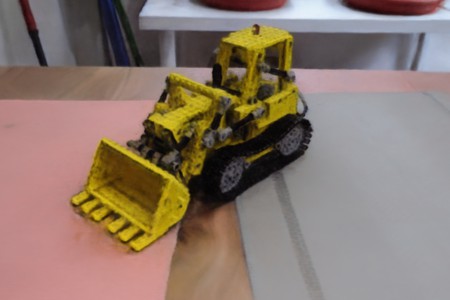} & \includegraphics[width=0.195\textwidth]{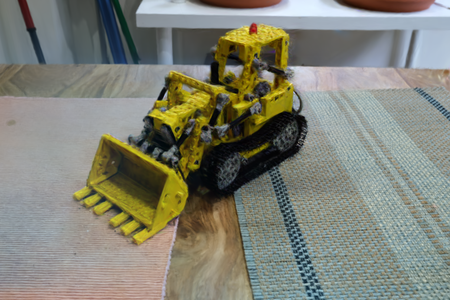} & \includegraphics[width=0.195\textwidth]{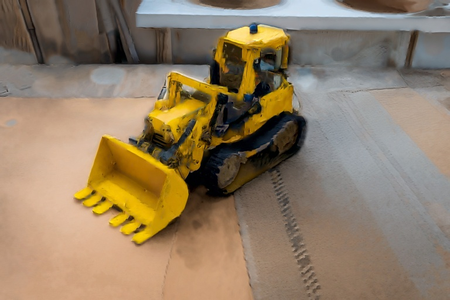} & \includegraphics[width=0.195\textwidth]{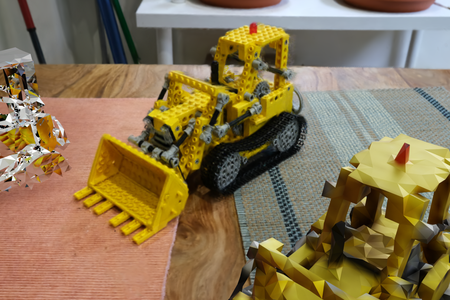} \\

       & Input & Instruct-N2N \cite{instructnerf2023} & DGE \cite{chen2024dge} & FRESCO \cite{yang2024fresco} & Ours \\
        
        \end{tabular}
    
    }
    \vspace{-3mm}
    \caption{{\bf Qualitative comparison on static editing.}}
    \vspace{-3mm}
    \label{fig:qual_static}
\end{figure*}

In Fig.~\ref{fig:qual_static}, we compare the visual quality of static scene editing across different methods. Our approach outperforms baselines in object insertion and manipulation, delivering realistic and accurate edits while preserving scene structure. In contrast, Instruct-N2N struggles with localized editing, FRESCO fails at structural preservation, and DGE, while producing realistic videos, cannot ensure instruction alignment. Additionally, AutoVFX provides richer capabilities, such as precise material editing (“{\it make it mirror-like}”), accurate object counting (“{\it drop five basketballs}”), and advanced visual effects (“{\it make it on fire}”).

\begin{figure*}[t]
    \centering
    \setlength\tabcolsep{0.1em}
    {%
    
        \begin{tabular}{ccccc}

        \textbf{Input} & \multicolumn{4}{c}{\small\textit{\makecell{``Insert a physics-enabled Benz 20 meters in front of us with random 2D rotation.\\ Add a Ferrari moving forward.''} }} \\
        
       \includegraphics[width=0.195\textwidth]{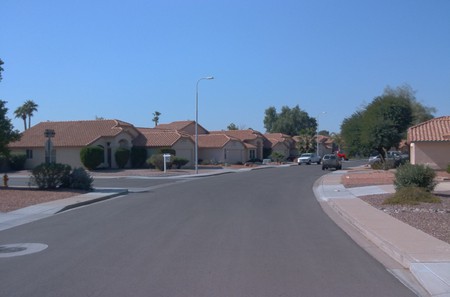} &
       \includegraphics[width=0.195\textwidth]{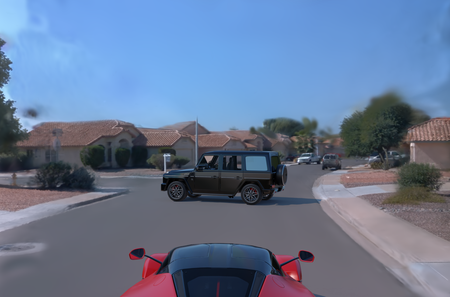} & \includegraphics[width=0.195\textwidth]{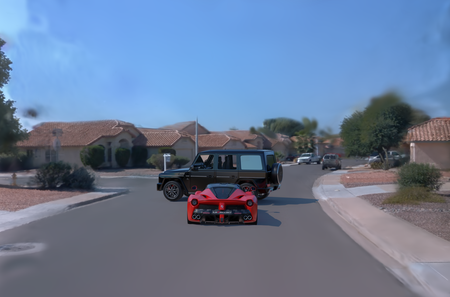} & \includegraphics[width=0.195\textwidth]{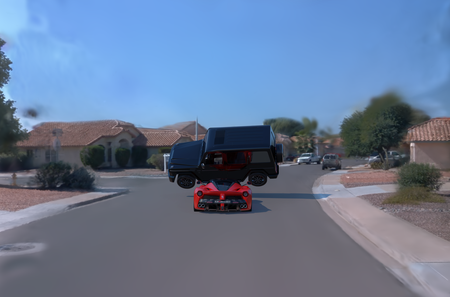} & 
       \includegraphics[width=0.195\textwidth]{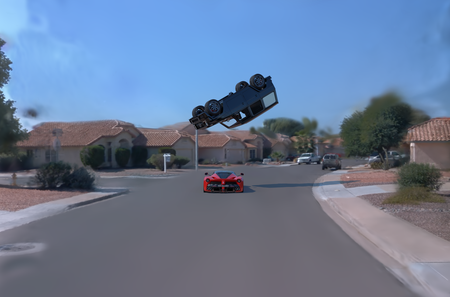} \\
       
        \end{tabular}
    }
    \vspace{-3mm}
    \caption{{\bf Dynamic simulation of \method on driving scenes.}}
    \vspace{-5mm}
    \label{fig:qual_waymo_ours}
\end{figure*}

\paragraph{Dynamic video simulation}
We present additional results for dynamic VFX video in Fig.~\ref{fig:qual_extra}. These highlight our method's ability to generate a wide range of realistic, physically plausible dynamic simulations from text instructions, using modules like rigid body simulations, object animation, smoke and fire, and object fracturing. None of the generative editing baselines support this feature. We also conduct experiments on autonomous driving simulation using the Waymo dataset~\cite{sun2020scalability}, as shown in Fig.~\ref{fig:qual_waymo_ours}. AutoVFX enables both realistic rendering and realistic physical interaction between cars in collision scenarios.

\subsection{Quantitative evaluation}
We also provide a quantitative evaluation of our method. The evaluation is based on nine metrics categorized into three groups: \textit{``Semantic Consistency Measures''}, \textit{``Multimodal LLM Quality Evaluation''} and \textit{``User Study''}. These metrics collectively provide a comprehensive assessment of the quality and effectiveness of text-guided visual edits. The quantitative results are presented in Table~\ref{tab:quant_all}.

\paragraph{Semantic Consistency Measures}
We incorporate the ``Multiple Objects'' metric from VBench~\cite{huang2023vbench} to verify the presence of objects after editing. This metric assesses whether multiple objects are correctly composed within the edits using a detection module, ensuring that the desired semantic content has been successfully modified. Instead of using GRiT~\cite{wu2022grit} as the detection module, we employ Grounded-SAM~\cite{ren2024grounded} for this task. We assess the success rate of visual content editing across all frames and for all possible edits. We also adopt ``CLIP Similarity'' and ``CLIP Directional Similarity'' metrics as proposed in DGE~\cite{chen2024dge} for evaluation. ``CLIP Similarity'' measures the alignment between the text instructions and each edited frame, while ``CLIP Directional Similariy'' evaluates the temporal consistency of the edits across frames. Both metrics operate in CLIP space. As shown in Table~\ref{tab:quant_all}, our method significantly outperforms other approaches in object detection score and CLIP directional similarity score, while achieving comparable results in CLIP similarity score. In particular, we improve the object detection score by a large margin, suggesting that our video edits reflect the goal of object-level changes. We notice that CLIP similarity is less discriminative among all methods and conjecture that this might be because global CLIP is not sensitive to capturing local changes, such as the insertion of small objects or dynamics. These results indicate that our method effectively aligns the edited outcomes with the provided text instructions.

\paragraph{Multimodal LLM Quality Evaluation}
Inspired by \cite{wu2024comprehensive}, we utilize multimodal LLMs as a powerful, interpretable, text-driven model for image quality assessment. Specifically, we prompt GPT-4o to evaluate and compare the ``Overall Perceptual Quality'' of four different methods based on three criteria: ``Text Alignment'', ``Photorealism'', and ``Structural Preservation''. We also ask GPT-4o to assign a quality score of each criterion to each method, ranging from 0 to 1, with 1 being the highest (detailed prompts could be found in supplementary). From Table~\ref{tab:quant_all}, our method outperforms other approaches across all four metrics by a significant margin. In particular, \method~creates more realistic videos, preserves structure better, and excels in ``Text Alignment'' by a most prominent margin. This demonstrates that our video editing produce high-quality and reasonable image edits that fully reflect the desired text instructions, which is desirable in downstream VFX applications.

\paragraph{User Study}
We conduct a user study to evaluate ``Text Alignment'' and ``Overall Realism'' of performed edits. To address the potential bias where minimal changes to visual content are perceived as more realistic, we structured the survey as follows: first, users are asked to evaluate which edited videos best aligned with the given text instructions, allowing for multiple selections. In the second question, users are required to choose the most realistic video from the set of choices they previously selected. We collected a total of 36 user samples. For detailed information on the user study methodology, please refer to the supplementary materials. As shown in Table~\ref{tab:quant_all}, our method receives a higher preference from users in both "Text Alignment" and "Overall Realism" categories, showing that the edits are not only accurate but also appealing to human judgment.

\section{Conclusion}
We presented \method{}, a system that automatically creates physically-grounded VFX given a monocular video and natural language instructions. \method{} combines neural scene modeling, LLM-based code generation, and physical simulation to allow realistic and easily controllable VFX creation. Experimental results demonstrate that \method{} outperforms existing scene editing methods based on a variety of practical criteria. We envision that \method{} will facilitate both the acceleration and democratization of visual content creation, helping both experienced artists and everyday users create the high-quality VFX that they desire.

\paragraph{Acknowledgement} This project is supported by the Intel AI SRS gift, Meta research grant, the IBM IIDAI Grant and NSF Awards \#2331878, \#2340254, \#2312102, \#2414227, and \#2404385. Hao-Yu Hsu is supported by Siebel Scholarship. We greatly appreciate the NCSA for providing computing resources. We thank Derek Hoiem, Sarita Adve, Benjamin Ummenhofer, Kai Yuan, Micheal Paulitsch, Katelyn Gao, Quentin Leboutet for helpful discussions.

{
    \small
    \bibliographystyle{ieeenat_fullname}
    \bibliography{main}
}

\clearpage
\setcounter{page}{1}
\maketitlesupplementary

\begin{figure*}[ht]
    \centering
    \setlength\tabcolsep{0.1em}
    \resizebox{1.0\textwidth}{!}{%
        \begin{tabular}{@{}cccccc@{}}

        \multicolumn{3}{c}{\prompt{Drop 5 basketballs on the table.}} & \multicolumn{3}{c}{\prompt{Insert an animated Goku figurine on the ground and make it on fire.}} \\
        
        \includegraphics[width=0.28\textwidth]{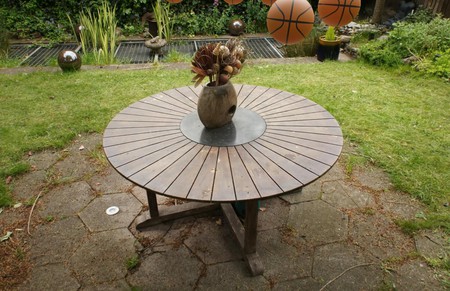} & \includegraphics[width=0.28\textwidth]{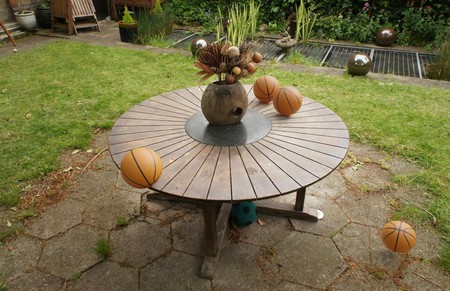} & \includegraphics[width=0.28\textwidth]{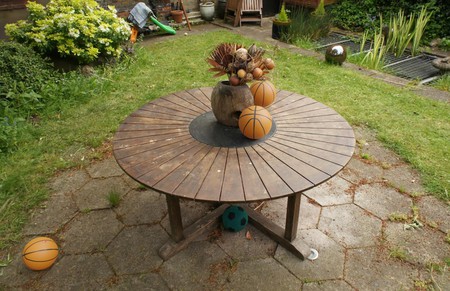} & \includegraphics[width=0.28\textwidth]{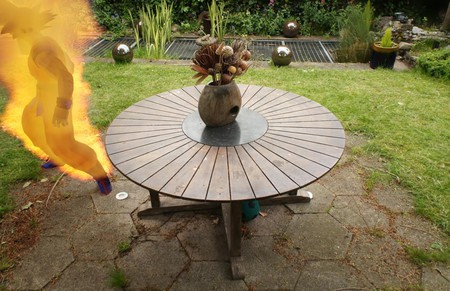} & \includegraphics[width=0.28\textwidth]{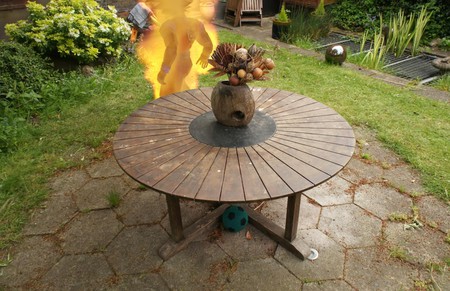} & \includegraphics[width=0.28\textwidth]{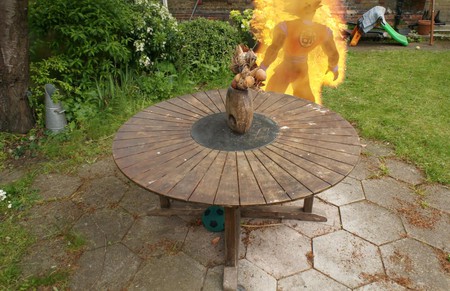} \\
       
        \multicolumn{3}{c}{\prompt{Place a pikachu with pebble material and a pikachu with rosewood material on the table.}} & \multicolumn{3}{c}{\prompt{Insert a red pikachu and a cyan pikachu on the table.}} \\
        
        \includegraphics[width=0.28\textwidth]{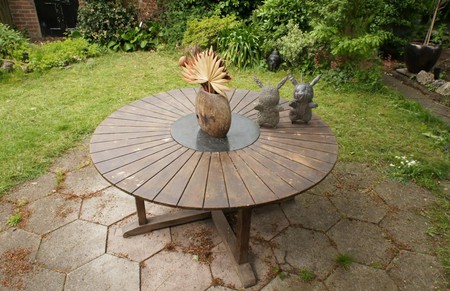} & \includegraphics[width=0.28\textwidth]{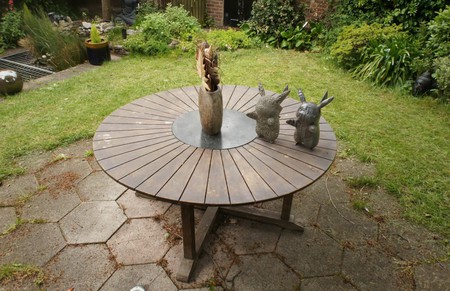} & \includegraphics[width=0.28\textwidth]{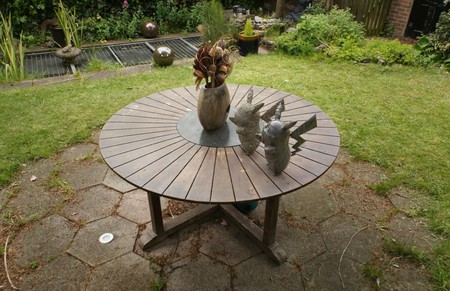} & \includegraphics[width=0.28\textwidth]{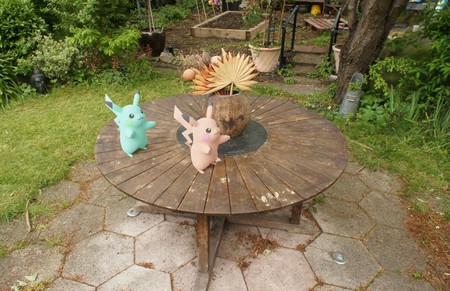} & \includegraphics[width=0.28\textwidth]{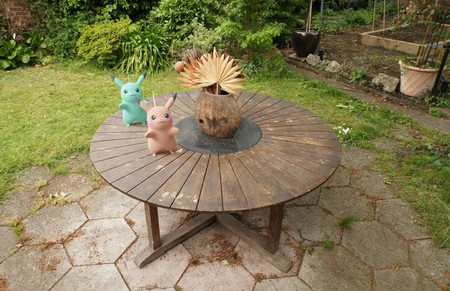} & \includegraphics[width=0.28\textwidth]{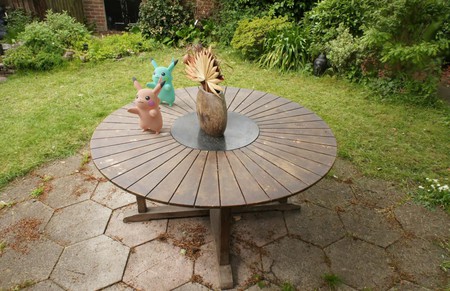} \\

        \multicolumn{3}{c}{\prompt{Make a bird flying around and above the table.}} & \multicolumn{3}{c}{\prompt{Generate a zombie character wearing hat and put it on the table.}} \\
        
        \includegraphics[width=0.28\textwidth]{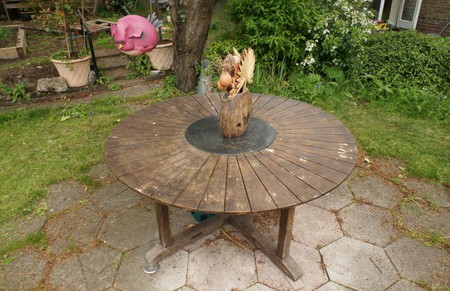} & \includegraphics[width=0.28\textwidth]{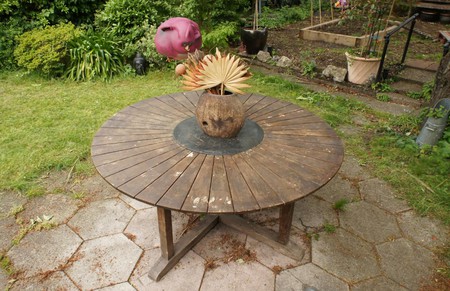} & \includegraphics[width=0.28\textwidth]{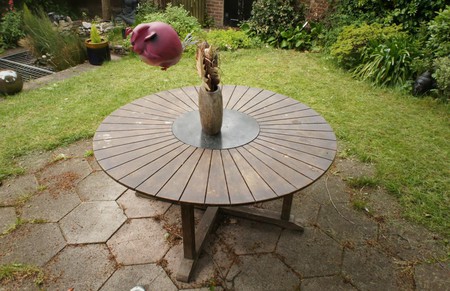} & \includegraphics[width=0.28\textwidth]{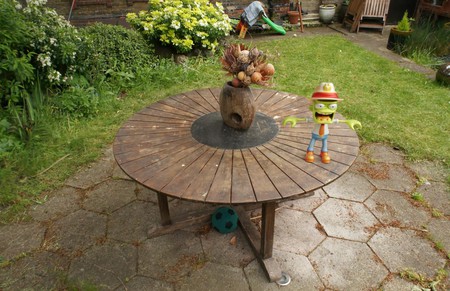} & \includegraphics[width=0.28\textwidth]{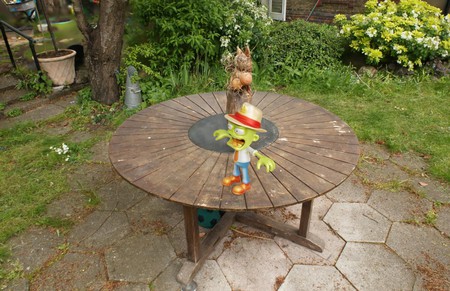} & \includegraphics[width=0.28\textwidth]{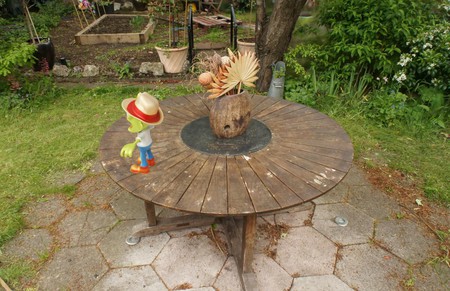} \\

        \multicolumn{3}{c}{\prompt{Make the vase with flowers 1.5 times bigger, then make it drop onto the table.}} & \multicolumn{3}{c}{\prompt{Rotate the vase with flowers by 90 degrees, then drop a basketball from the top of this vase to make it fracture.}} \\
        
        \includegraphics[width=0.28\textwidth]{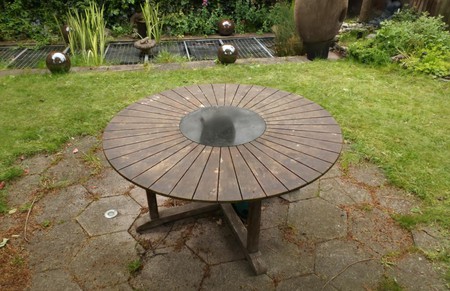} & \includegraphics[width=0.28\textwidth]{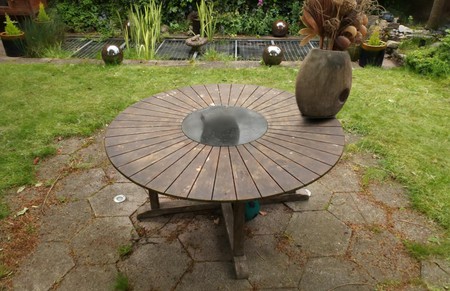} & \includegraphics[width=0.28\textwidth]{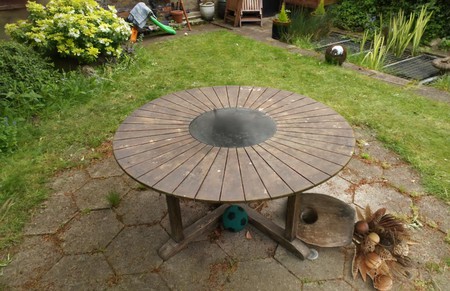} & \includegraphics[width=0.28\textwidth]{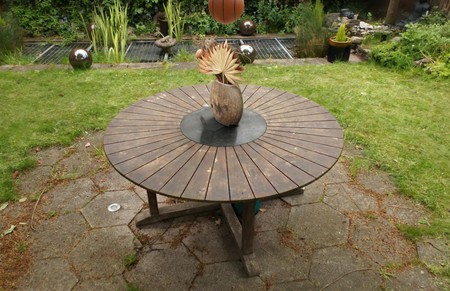} & \includegraphics[width=0.28\textwidth]{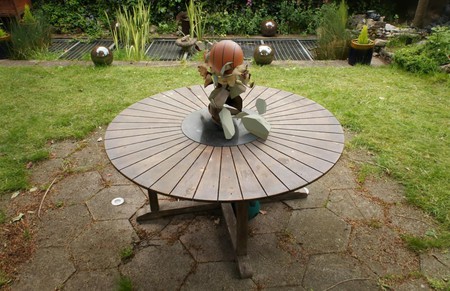} & \includegraphics[width=0.28\textwidth]{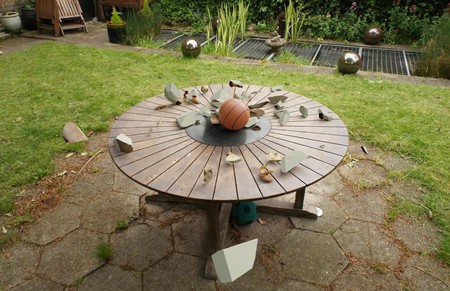} \\

        \multicolumn{3}{c}{\prompt{Place a trophy on the table.}} & \multicolumn{3}{c}{\prompt{Put a Tony Stark on the floor covered with smoke.}} \\
        
        \includegraphics[width=0.28\textwidth]{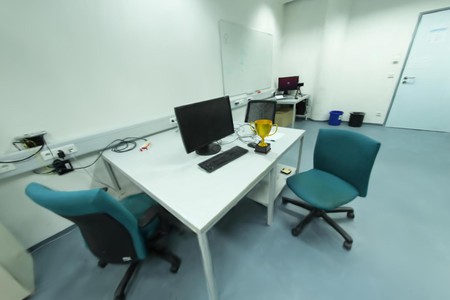} & \includegraphics[width=0.28\textwidth]{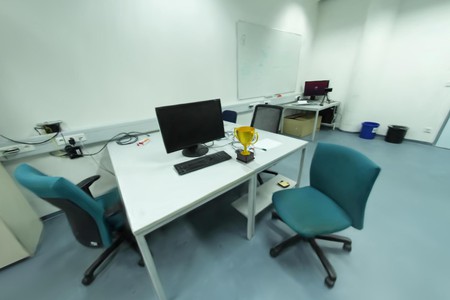} & \includegraphics[width=0.28\textwidth]{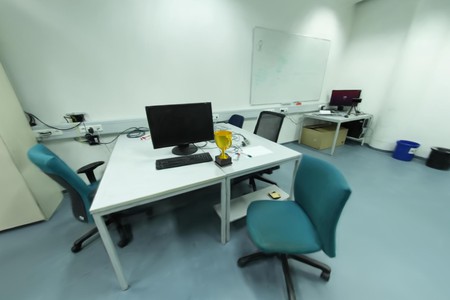} & \includegraphics[width=0.28\textwidth]{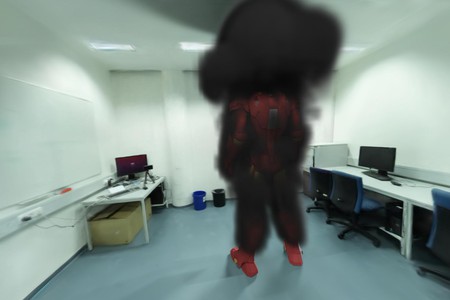} & \includegraphics[width=0.28\textwidth]{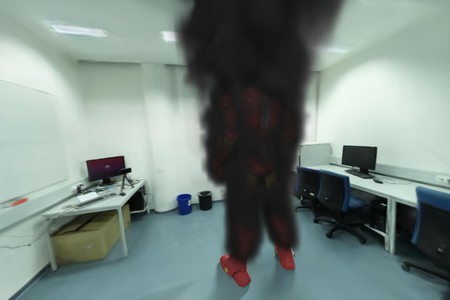} & \includegraphics[width=0.28\textwidth]{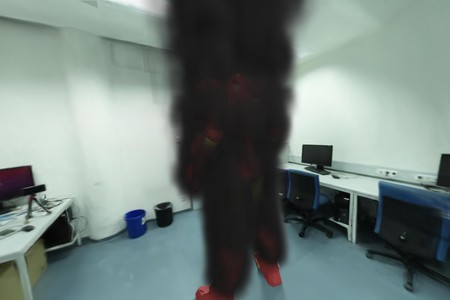} \\

        \multicolumn{3}{c}{\prompt{Drop 10 forks onto the blue gloves.}} & \multicolumn{3}{c}{\prompt{Insert a rusty chair on the carpet.}} \\
        
        \includegraphics[width=0.28\textwidth]{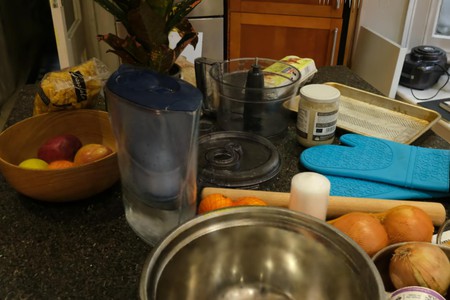} & \includegraphics[width=0.28\textwidth]{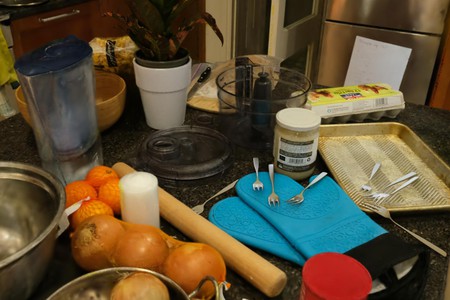} & \includegraphics[width=0.28\textwidth]{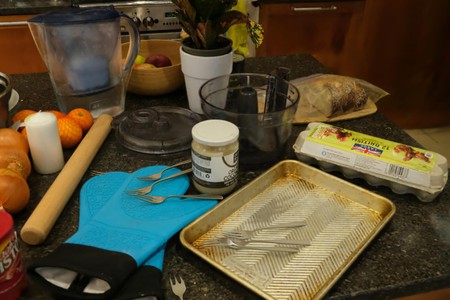} & \includegraphics[width=0.28\textwidth]{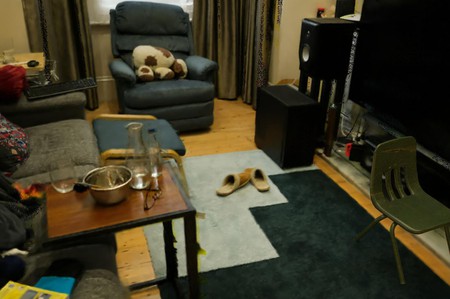} & \includegraphics[width=0.28\textwidth]{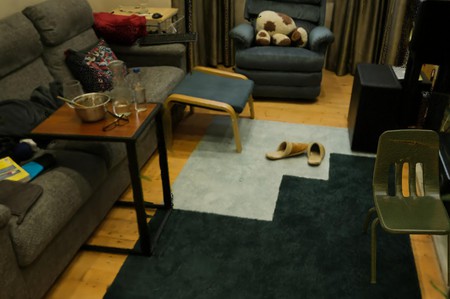} & \includegraphics[width=0.28\textwidth]{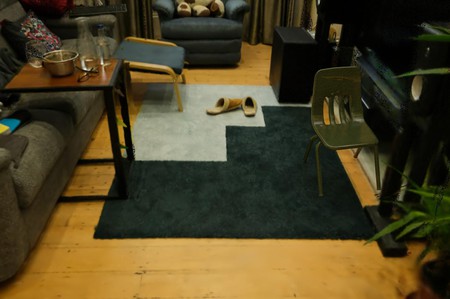} \\

        \multicolumn{3}{c}{\prompt{Break the sculpture.}} & \multicolumn{3}{c}{\prompt{Drop seven apples on the floor and paint them in rainbow colors.}} \\
        
        \includegraphics[width=0.28\textwidth]{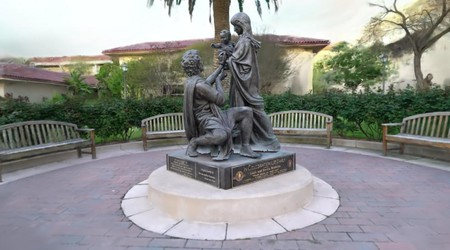} & \includegraphics[width=0.28\textwidth]{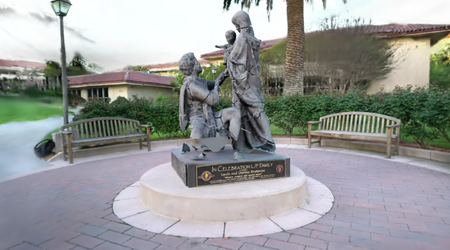} & \includegraphics[width=0.28\textwidth]{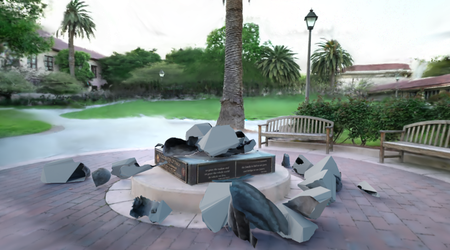} & \includegraphics[width=0.28\textwidth]{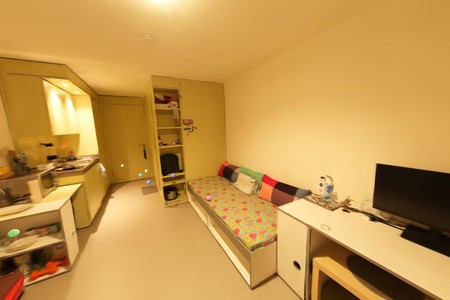} & \includegraphics[width=0.28\textwidth]{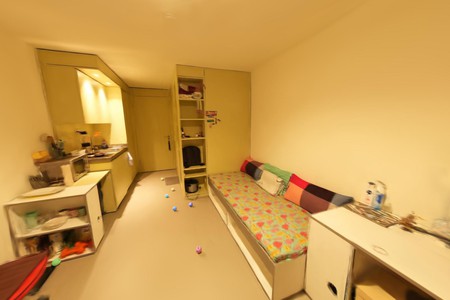} & \includegraphics[width=0.28\textwidth]{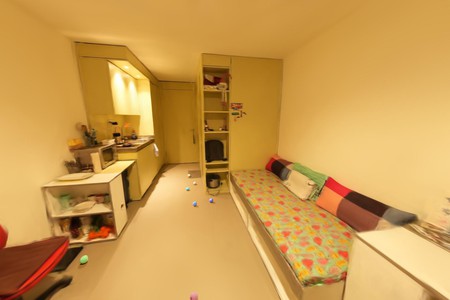} \\
       
    \end{tabular}    
    }
    \caption{More editing results using \method.}
    \label{fig:supp_qual_extra}
\end{figure*}

\begin{figure*}[ht]
    \centering\setlength{\tabcolsep}{2pt}
    \resizebox{1.0\textwidth}{!}{%
        \begin{tabular}{@{}ccccc@{}}

        {\Large\textbf{Input Observation}} & \multicolumn{4}{c}{\Large\textit{``Insert a physics-enabled Benz G 20 meters in front of us with random 2D rotation. Add a Ferriari moving forward.''}} \\

       \includegraphics[width=0.4\textwidth]{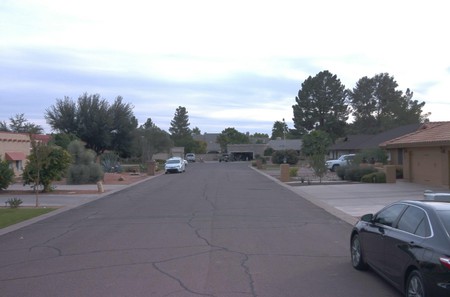} & \includegraphics[width=0.4\textwidth]{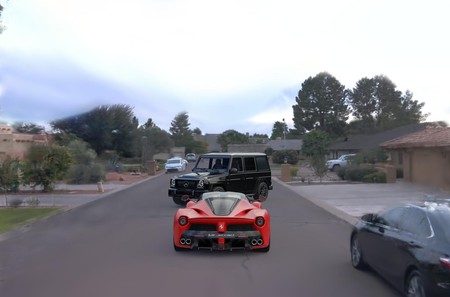} & \includegraphics[width=0.4\textwidth]{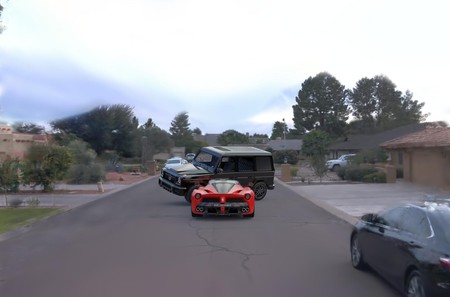} & \includegraphics[width=0.4\textwidth]{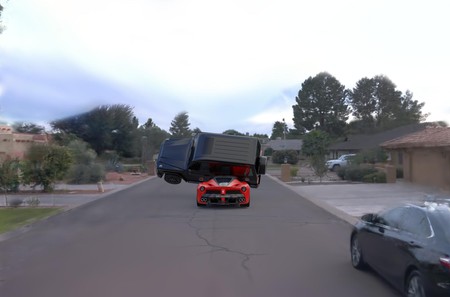} &
       \includegraphics[width=0.4\textwidth]{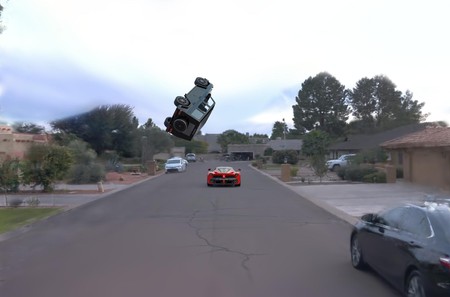} \\

        {\Large\textbf{Input Observation}} & \multicolumn{4}{c}{\Large\textit{``Insert a statue to be able to fracture 20 meters in front of our vehicle. Then, make a Porsche driving forward.''}} \\

       \includegraphics[width=0.4\textwidth]{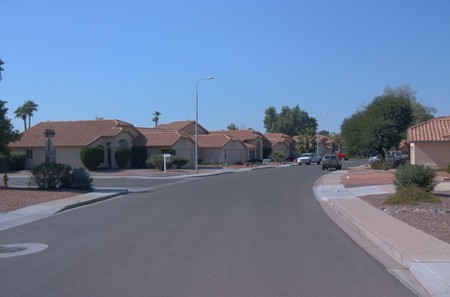} & \includegraphics[width=0.4\textwidth]{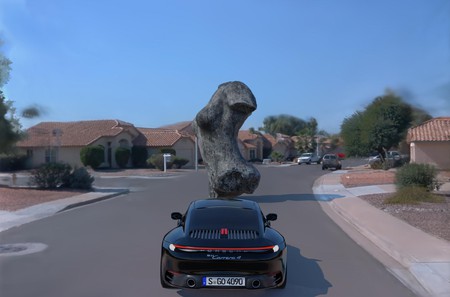} & \includegraphics[width=0.4\textwidth]{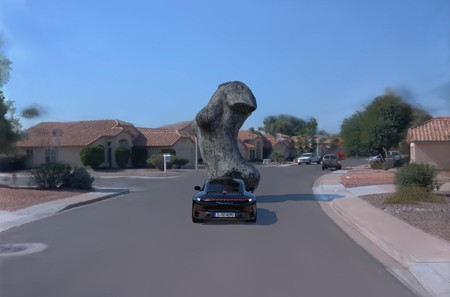} & \includegraphics[width=0.4\textwidth]{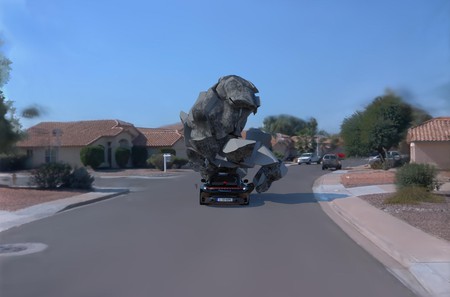} &
       \includegraphics[width=0.4\textwidth]{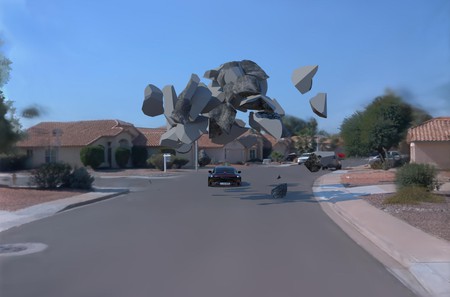} \\

        {\Large\textbf{Input Observation}} & \multicolumn{4}{c}{\Large\textit{``Drop a Benz G with fire randomly in 10 meters front of our vehicles and from 3 meters high.''}} \\

       \includegraphics[width=0.4\textwidth]{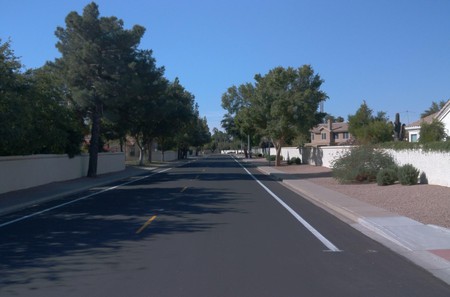} & \includegraphics[width=0.4\textwidth]{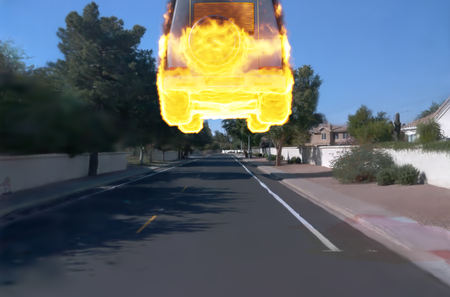} & \includegraphics[width=0.4\textwidth]{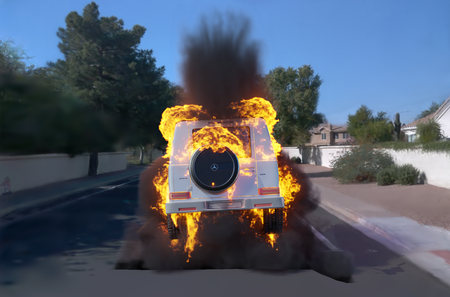} & \includegraphics[width=0.4\textwidth]{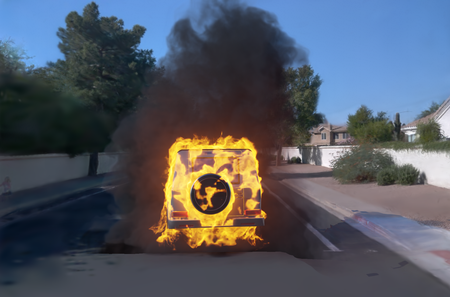} &
       \includegraphics[width=0.4\textwidth]{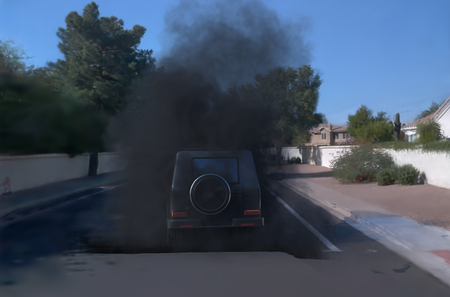} \\

        \end{tabular}
        
    }
    \caption{More dynamic simulation results of \method{} on autonomous driving scenes.}
    \label{fig:supp_qual_chatsim_ours}
\end{figure*}

\section{Implementation details}
In this section, we provide an overview of our framework, followed by a detailed explanation of the implementation, including scene modeling, simulation, rendering, composition, and LLM integration. We plan to release the entire codebase upon acceptance.

\subsection{Holistic overview}
We use Blender's modules~\cite{blender} to implement all the editing, simulation, and rendering components. These include Cycles renderer, Material Nodes, Mantaflow fluid simulation and Composition Nodes. We chose Blender because: (1) it includes all the necessary modules required by AutoVFX, and (2) it offers a convenient Python-based interface for modular function encapsulation and code generation. However, AutoVFX is generic, allowing the easy integration of new modules for additional functionality. One can choose different low-level implementations, whether Blender or other tools with a Python-based interface, such as Mitsuba for rendering~\cite{Mitsuba3} or Taichi for simulation~\cite{hu2019taichi, hu2019difftaichi, hu2021quantaichi}.

\subsection{Scene modeling details}

\paragraph{Geometry}
We employ BakedSDF~\cite{yariv2023bakedsdf}, implemented in SDFStudio~\cite{Yu2022SDFStudio}, to obtain high-quality scene geometry due to its detailed mesh extraction. Specifically, we use \textit{bakedsdf-mlp} model. This model is trained for 250k steps using default optimization and model settings, with an additional monocular normal consistency loss set by \textit{pipeline.model.mono-normal-loss-mult=0.1}. Monocular normal maps are obtained from Omnidata~\cite{eftekhar2021omnidata}. For fully-captured indoor scenes such as ScanNet++~\cite{yeshwanth2023scannet++}, we enable the inside-outside flag with \textit{pipeline.model.sdf-field.inside-outside=True}. For scenes with distant backgrounds, we enable background modeling by setting \textit{pipeline.model.background-model=mlp}.

While BakedSDF excels in capturing object-centric scenes, it struggles with non-object-centric, long, and narrow camera trajectories, such as those in street views for autonomous driving. To address this limitation, we use StreetSurf~\cite{guo2023streetsurf} for geometry reconstruction in road scenes from the Waymo dataset~\cite{sun2020scalability}. For a fair comparison, we do not utilize LiDAR point clouds for precise geometry initialization; instead, we use three camera views (Front, Front Left, Front Right), consistent with ChatSim~\cite{wei2024editable}, along with monocular normal and depth priors from Omnidata, and sky masks extracted using SegFormer~\cite{xie2021segformer}.

\paragraph{Appearance \& Semantics}
To model appearance, we use both 3D Gaussian Splatting~\cite{kerbl20233d} and SuGaR~\cite{guedon2023sugar}. The model is first trained with 3D Gaussian Splatting for 15000 steps, followed by an additional 7000 steps using SuGaR, all with default optimization parameters. To achieve a denser initialization for better rendering quality, we enhance the Gaussian initialization from COLMAP~\cite{schoenberger2016sfm} points by computing ray-mesh intersections for each training view and assigning pixel RGB values and intersected points to set up the Gaussians. For loss terms, we apply the anisotropic regularizer from PhysGaussian~\cite{xie2023physgaussian} to prevent the emergence of spiky Gaussians during training. Additionally, we incorporate normal regularization from GaussianShader~\cite{jiang2024gaussianshader} to ensure consistency between local geometry and estimated normals. An anisotropic loss weight of 0.1 and a normal loss weight of 0.01 are used across all scenes. These regularizations help maintain the Gaussians' shape and orientation, facilitating better instance extraction. To avoid false-positive predictions in the semantic branches, we increase the DINO~\cite{caron2021emerging} threshold to 0.45 in DEVA~\cite{cheng2023tracking}. Full pseudo code for 3D instance segmentation on both meshes and 3D Gaussians are illustrated in Fig.~\ref{fig:supp_extract_algo}.

\paragraph{Lighting}
To illuminate the scene with surrounding light, we extract an HDR environmental map from a single image using DiffusionLight~\cite{phongthawee2023diffusionlight}. We begin by center-cropping the image to 512x512 pixels, then inpainting a chrome ball using a diffusion model. The chrome ball is subsequently unwrapped to create the environmental map. Multiple chrome balls with varying exposure values are generated and merged to produce the final HDR map. This map is then transformed based on the camera poses of the original image to align it with the world space.

\begin{figure}[t]
    \centering\setlength{\tabcolsep}{2pt}
    \resizebox{0.45\textwidth}{!}{%
        \begin{tabular}{@{}cc@{}}

        \Large\textbf{Without sunlight} & \Large\textbf{With sunlight} \\

        \includegraphics[width=0.5\textwidth]{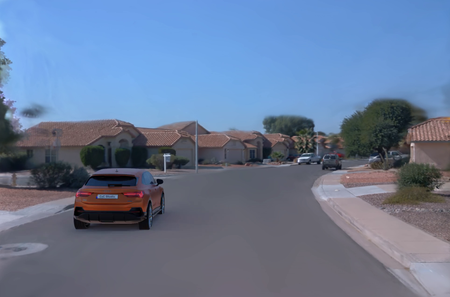} & \includegraphics[width=0.5\textwidth]{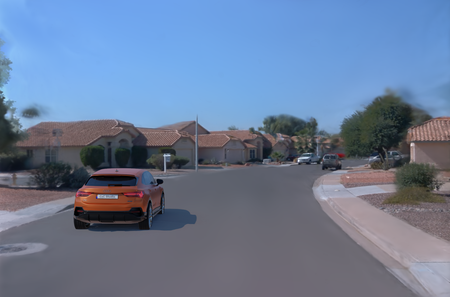} \\

        \end{tabular}
    }
    \caption{Comparison of simulation results with and without sunlight in Waymo scenes.}
    \label{fig:supp_chatsim_diff}
\end{figure}

For consistent lighting effects in Blender, we adjust the HDR map's intensity according to the scene type: 0.6 for outdoor scenes and 2.0 for indoor scenes. In fully-captured indoor environments like ScanNet++~\cite{yeshwanth2023scannet++}, where HDR maps are insufficient due to occlusions by surrounding geometry like walls and ceilings, we extract emitter meshes by unprojecting over-saturated pixels into 3D space and using majority voting to estimate the emitter locations. These meshes are imported into Blender as white-colored emitters, with their strength set to 100. For outdoor autonomous driving scenes, such as those in Waymo~\cite{sun2020scalability}, the HDR map alone is insufficient for casting strong shadows. To address this, we determine the sunlight direction from the brightest area in the HDR map and add a corresponding sunlight source in Blender, enhancing shadow realism on the road. The impact of this additional sunlight source is illustrated in Fig.~\ref{fig:supp_chatsim_diff}.

\subsection{Scene simulation details}

\paragraph{Animation and rigidbody simulation}
To simulate the movement of animated objects along a series of 3D keypoints, we use Bézier curves~\cite{prautzsch2002bezier} to generate a smooth, continuous path from discrete sample positions, ensuring seamless transitions of the animated objects. We additionally model object-scene rigid body interactions using Blender, which is based on the Bullet physics engine~\cite{coumans2015bullet}. To achieve both accurate and realistic interactions, we also pre-compute the center of mass and convex hull for collision checking of any interactive objects.
For object assets extracted from the scene, which require rendering with 3D Gaussians post-simulation, we preserve the rigid body transformations at each timestep. These transformations are then applied to the 3D Gaussians during rendering. This process closely follows the principles of recent works on deforming Gaussians~\cite{qiu-2024-featuresplatting, xie2023physgaussian, wen2024gomavatar}, where the centroids and covariance of the Gaussians are adjusted through translation, rotation, and scaling.

\paragraph{Physical effects}
Realistic VFX effects often require compelling physical simulations, such as fracture effects or particle effects like smoke and fire. For fracture effects, we employ the cell fracture algorithm~\cite{cellfracture} to generate self-fracturing objects. We configure the fracture count to 100 and apply the object's average color to the internal fractures. For particle effects, we adopt Blender's computational fluid dynamics addon Mantaflow~\cite{mantaflow}, which is an efficient implementation of the FLIP-based~\cite{brackbill1988flip} particle simulation method, to simulate smoke emission. To balance computational efficiency and quality in Blender, we configure the smoke domain with a resolution of 128, an adaptive margin of 4, an adaptive threshold of 0.005, and a dissolve speed of 30. We modify the material nodes to further enhance the realism of smoke and fire effects. For smoke simulation, we set the smoke color to (0.1, 0.1, 0.1, 1) and the smoke density to 70. For fire simulation, we reduce the smoke density to 50, set the object’s temperature to 1500, and configure the blackbody tint and intensity to (1, 0.3886, 0.0094, 1) and 5, respectively.

\subsection{Rendering \& composition details}

\paragraph{Rendering}
We use Blender's Cycles renderer for rendering. Cycles is Blender's physically-based path tracing renderer, designed for high-quality, photorealistic rendering. It accurately simulates light interactions, including reflections, refractions, and global illumination, making it ideal for realistic visual effects and animations. In our workflow, we render three outputs: foreground objects, background meshes, and a combined render of the two, as detailed in the main paper. To make foreground objects affected by lighting from the background, we set \texttt{visible\_camera=False} for background meshes to make them invisible to the camera on the first light bounce but still affects subsequent bounces. The default number of samples in Cycles are set to 64, increased to 512 for scenes involving smoke and fire simulations to better capture particle effect details. Images are rendered at 2x resolution to mitigate aliasing during compositing. 

\begin{figure*}[t]
    \centering
    \includegraphics[width=1.0\linewidth]{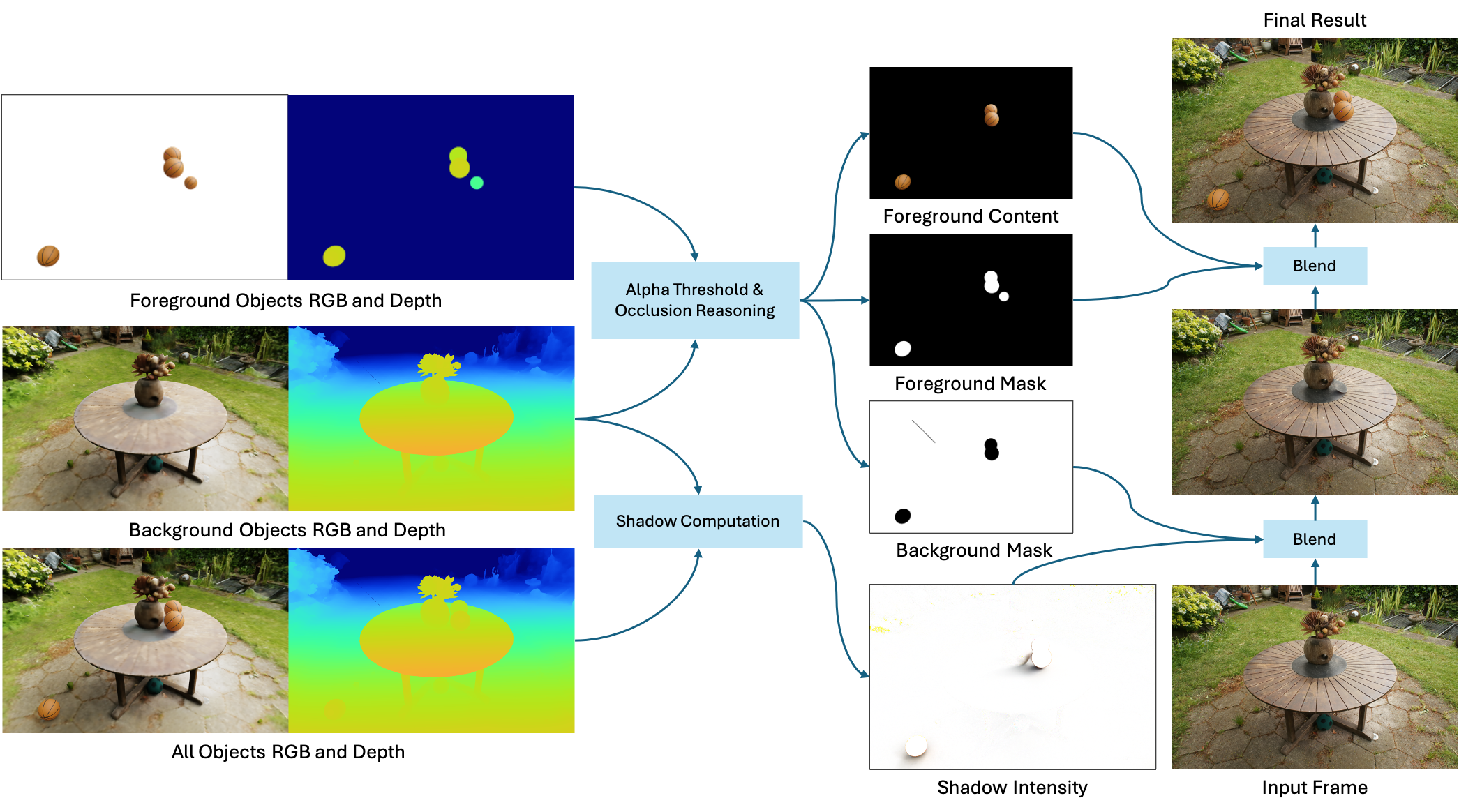}
    \caption{Our image composition pipeline. 
The process starts by generating foreground and background masks, along with foreground content, through alpha thresholding and occlusion reasoning based on rendered objects and background meshes. Next, shadow intensity is calculated by determining the ratio of pixel values between the combined rendering of all objects and the background meshes. Finally, the shadows and foreground content are sequentially blended into the original image to produce the final result.}
    \label{fig:supp_composite_demo}
\end{figure*}

\paragraph{Compositing}
The final visual effects are achieved through a compositing pipeline that blends visual content into the original frames. This process involves extracting foreground and background masks, and foreground content via alpha thresholding and occlusion reasoning, and calculating shadow intensity as the pixel value ratio between the combined and background renders. Shadows are then blended into the original image, followed by the integration of foreground content, resulting in the final composited video. The composition pipeline is illustrated in Fig.~\ref{fig:supp_composite_demo}.

\subsection{LLM integration details}

\paragraph{Modular functions design}
The predefined editing modules are encapsulated into callable and executable functions that can be utilized by LLM. We provide a list of all designed modules, along with a brief introduction to each, including its purpose, inputs, and outputs. For further details on the editing modules, please refer to the attached file \texttt{edit\_utils.py}.
Details of editing modules:

\begin{itemize}
    \item \textbf{\texttt{detect\_object}}
    \begin{itemize}
        \item \textbf{Purpose:} Detects and extracts instance-level meshes from a scene.
        \item \textbf{Input:} 
        \begin{itemize}
            \item scene\_representation: The representation of the scene in which to detect the object.
            \item object\_name: The name of the object to be detected in the scene.
        \end{itemize}
        \item \textbf{Output:} 
        \begin{itemize}
            \item A dictionary containing information about the detected object.
        \end{itemize}
    \end{itemize}

    \item \textbf{\texttt{sample\_point\_on\_object}}
    \begin{itemize}
        \item \textbf{Purpose:} Samples a point on the surface of an object mesh.
        \item \textbf{Input:} 
        \begin{itemize}
            \item scene\_representation: The scene in which the object is located.
            \item obj: The object on which to sample a point.
        \end{itemize}
        \item \textbf{Output:} 
        \begin{itemize}
            \item A 3D point location on the object.
        \end{itemize}
    \end{itemize}

    \item \textbf{\texttt{sample\_point\_above\_object}}
    \begin{itemize}
        \item \textbf{Purpose:} Samples a point above an object at a specified vertical offset.
        \item \textbf{Input:} 
        \begin{itemize}
            \item scene\_representation: The scene in which the object is located.
            \item obj: The object above which to sample a point.
            \item VERTICAL\_OFFSET: The vertical distance above the object to sample the point (optional).
        \end{itemize}
        \item \textbf{Output:} 
        \begin{itemize}
            \item A 3D point location above the object.
        \end{itemize}
    \end{itemize}

    \item \textbf{\texttt{retrieve\_asset}}
    \begin{itemize}
        \item \textbf{Purpose:} Retrieves a 3D asset by its name from objaverse.
        \item \textbf{Input:} 
        \begin{itemize}
            \item scene\_representation: The scene in which to retrieve the asset.
            \item object\_name: The name of the asset to retrieve.
            \item is\_animated: Boolean flag indicating if the asset is animated (optional).
            \item is\_generated: Boolean flag indicating if the asset is generated (optional).
        \end{itemize}
        \item \textbf{Output:} 
        \begin{itemize}
            \item A dictionary containing information about the retrieved object.
        \end{itemize}
    \end{itemize}

    \item \textbf{\texttt{insert\_object}}
    \begin{itemize}
        \item \textbf{Purpose:} Inserts an object into the scene.
        \item \textbf{Input:} 
        \begin{itemize}
            \item scene\_representation: The scene representation into which the object is inserted.
            \item obj: The object to insert into the scene.
        \end{itemize}
        \item \textbf{Output:} None
    \end{itemize}

    \item \textbf{\texttt{remove\_object}}
    \begin{itemize}
        \item \textbf{Purpose:} Removes an object from the scene, with optional inpainting.
        \item \textbf{Input:} 
        \begin{itemize}
            \item scene\_representation: The scene from which the object is to be removed.
            \item obj: The object to be removed.
            \item remove\_gaussians: Boolean flag to determine if associated Gaussian splatting should also be removed (optional).
        \end{itemize}
        \item \textbf{Output:} None
    \end{itemize}

    \item \textbf{\texttt{update\_object}}
    \begin{itemize}
        \item \textbf{Purpose:} Updates an object's information in the scene.
        \item \textbf{Input:} 
        \begin{itemize}
            \item scene\_representation: The scene representation that contains the object.
            \item obj: The object whose information is to be updated.
        \end{itemize}
        \item \textbf{Output:} None
    \end{itemize}

    \item \textbf{\texttt{allow\_physics}}
    \begin{itemize}
        \item \textbf{Purpose:} Enables rigid body simulation for an object.
        \item \textbf{Input:} 
        \begin{itemize}
            \item obj: The object to enable physics for.
        \end{itemize}
        \item \textbf{Output:} 
        \begin{itemize}
            \item Updated object dictionary with rigid body settings.
        \end{itemize}
    \end{itemize}

    \item \textbf{\texttt{add\_fire}}
    \begin{itemize}
        \item \textbf{Purpose:} Adds fire to an object in the scene.
        \item \textbf{Input:} 
        \begin{itemize}
            \item scene\_representation: The scene representation containing the object.
            \item obj: The object to which fire is added.
        \end{itemize}
        \item \textbf{Output:} None
    \end{itemize}

    \item \textbf{\texttt{add\_smoke}}
    \begin{itemize}
        \item \textbf{Purpose:} Adds smoke to an object in the scene.
        \item \textbf{Input:} 
        \begin{itemize}
            \item scene\_representation: The scene representation containing the object.
            \item obj: The object to which smoke is added.
        \end{itemize}
        \item \textbf{Output:} None
    \end{itemize}

    \item \textbf{\texttt{set\_static\_animation}}
    \begin{itemize}
        \item \textbf{Purpose:} Sets an object's animation to be static.
        \item \textbf{Input:} 
        \begin{itemize}
            \item obj: The object to set as static.
        \end{itemize}
        \item \textbf{Output:} 
        \begin{itemize}
            \item Updated object dictionary with animation settings.
        \end{itemize}
    \end{itemize}

    \item \textbf{\texttt{set\_moving\_animation}}
    \begin{itemize}
        \item \textbf{Purpose:} Sets an object's trajectory based on a list of 3D points.
        \item \textbf{Input:} 
        \begin{itemize}
            \item obj: The object to animate.
            \item points: List of 3D points defining the trajectory.
        \end{itemize}
        \item \textbf{Output:} 
        \begin{itemize}
            \item Updated object dictionary with trajectory settings.
        \end{itemize}
    \end{itemize}

    \item \textbf{\texttt{init\_material}}
    \begin{itemize}
        \item \textbf{Purpose:} Initializes a material instance with default values.
        \item \textbf{Input:} None
        \item \textbf{Output:} 
        \begin{itemize}
            \item An instance of the Material class.
        \end{itemize}
    \end{itemize}

    \item \textbf{\texttt{retrieve\_material}}
    \begin{itemize}
        \item \textbf{Purpose:} Retrieves a material by its name from PolyHaven.
        \item \textbf{Input:} 
        \begin{itemize}
            \item scene\_representation: The scene representation that requires the material.
            \item material\_name: The name of the material to retrieve.
        \end{itemize}
        \item \textbf{Output:} 
        \begin{itemize}
            \item Path to the material folder.
        \end{itemize}
    \end{itemize}

    \item \textbf{\texttt{apply\_material}}
    \begin{itemize}
        \item \textbf{Purpose:} Applies a material to an object.
        \item \textbf{Input:} 
        \begin{itemize}
            \item obj: The object to which the material is applied.
            \item material: The material instance to apply.
        \end{itemize}
        \item \textbf{Output:} 
        \begin{itemize}
            \item Updated object dictionary with applied material.
        \end{itemize}
    \end{itemize}

    \item \textbf{\texttt{allow\_fracture}}
    \begin{itemize}
        \item \textbf{Purpose:} Enables fracturing of an object.
        \item \textbf{Input:} 
        \begin{itemize}
            \item obj: The object to enable fracturing for.
        \end{itemize}
        \item \textbf{Output:} 
        \begin{itemize}
            \item Updated object dictionary with fracture settings.
        \end{itemize}
    \end{itemize}

    \item \textbf{\texttt{make\_break}}
    \begin{itemize}
        \item \textbf{Purpose:} Breaks an object into multiple pieces.
        \item \textbf{Input:} 
        \begin{itemize}
            \item obj: The object to break.
        \end{itemize}
        \item \textbf{Output:} 
        \begin{itemize}
            \item Updated object dictionary with break settings.
        \end{itemize}
    \end{itemize}

    \item \textbf{\texttt{make\_melting}}
    \begin{itemize}
        \item \textbf{Purpose:} Melts down an object into viscous liquid.
        \item \textbf{Input:} 
        \begin{itemize}
            \item obj: The object to melt down.
        \end{itemize}
        \item \textbf{Output:} 
        \begin{itemize}
            \item Updated object dictionary with melting settings.
        \end{itemize}
    \end{itemize}

    \item \textbf{\texttt{get\_object\_center\_position}}
    \begin{itemize}
        \item \textbf{Purpose:} Returns the position of the object at its center.
        \item \textbf{Input:} 
        \begin{itemize}
            \item obj: The object whose center position is required.
        \end{itemize}
        \item \textbf{Output:} 
        \begin{itemize}
            \item A 3D position vector.
        \end{itemize}
    \end{itemize}

    \item \textbf{\texttt{get\_object\_bottom\_position}}
    \begin{itemize}
        \item \textbf{Purpose:} Returns the position of the object at its bottom.
        \item \textbf{Input:} 
        \begin{itemize}
            \item obj: The object whose bottom position is required.
        \end{itemize}
        \item \textbf{Output:} 
        \begin{itemize}
            \item A 3D position vector.
        \end{itemize}
    \end{itemize}

    \item \textbf{\texttt{translate\_object}}
    \begin{itemize}
        \item \textbf{Purpose:} Translates an object by a given translation vector.
        \item \textbf{Input:} 
        \begin{itemize}
            \item obj: The object to translate.
            \item translation: The translation vector.
        \end{itemize}
        \item \textbf{Output:} 
        \begin{itemize}
            \item Updated object dictionary with new position.
        \end{itemize}
    \end{itemize}

    \item \textbf{\texttt{rotate\_object}}
    \begin{itemize}
        \item \textbf{Purpose:} Rotates an object by a given rotation matrix.
        \item \textbf{Input:} 
        \begin{itemize}
            \item obj: The object to rotate.
            \item rotation: The rotation matrix.
        \end{itemize}
        \item \textbf{Output:} 
        \begin{itemize}
            \item Updated object dictionary with new rotation.
        \end{itemize}
    \end{itemize}

    \item \textbf{\texttt{scale\_object}}
    \begin{itemize}
        \item \textbf{Purpose:} Scales an object by a given scale factor.
        \item \textbf{Input:} 
        \begin{itemize}
            \item obj: The object to scale.
            \item scale: The scale factor.
        \end{itemize}
        \item \textbf{Output:} 
        \begin{itemize}
            \item Updated object dictionary with new scale.
        \end{itemize}
    \end{itemize}

    \item \textbf{\texttt{get\_random\_2D\_rotation}}
    \begin{itemize}
        \item \textbf{Purpose:} Returns a random 2D rotation matrix (rotation around the z-axis).
        \item \textbf{Input:} None
        \item \textbf{Output:} 
        \begin{itemize}
            \item 3x3 rotation matrix.
        \end{itemize}
    \end{itemize}

    \item \textbf{\texttt{get\_random\_3D\_rotation}}
    \begin{itemize}
        \item \textbf{Purpose:} Returns a random 3D rotation matrix.
        \item \textbf{Input:} None
        \item \textbf{Output:} 
        \begin{itemize}
            \item 3x3 rotation matrix.
        \end{itemize}
    \end{itemize}

    \item \textbf{\texttt{make\_copy}}
    \begin{itemize}
        \item \textbf{Purpose:} Creates a deep copy of an object.
        \item \textbf{Input:} 
        \begin{itemize}
            \item obj: The object to copy.
        \end{itemize}
        \item \textbf{Output:} 
        \begin{itemize}
            \item New object dictionary with a unique object\_id.
        \end{itemize}
    \end{itemize}

    \item \textbf{\texttt{add\_event}}
    \begin{itemize}
        \item \textbf{Purpose:} Adds an event to the scene involving an object.
        \item \textbf{Input:} 
        \begin{itemize}
            \item scene\_representation: The scene representation to which the event is added.
            \item obj: The object involved in the event.
            \item event\_type: The type of event to add (e.g., "break", "incinerate").
            \item start\_frame: The frame at which the event starts (optional).
            \item end\_frame: The frame at which the event ends (optional).
        \end{itemize}
        \item \textbf{Output:} None
    \end{itemize}

    \item \textbf{\texttt{get\_camera\_position}}
    \begin{itemize}
        \item \textbf{Purpose:} Returns the camera position.
        \item \textbf{Input:} 
        \begin{itemize}
            \item scene\_representation: The scene representation containing the camera.
        \end{itemize}
        \item \textbf{Output:} 
        \begin{itemize}
            \item 3D position vector.
        \end{itemize}
    \end{itemize}

    \item \textbf{\texttt{get\_vehicle\_position}}
    \begin{itemize}
        \item \textbf{Purpose:} Returns the position of a vehicle in the scene.
        \item \textbf{Input:} 
        \begin{itemize}
            \item scene\_representation: The scene representation containing the vehicle.
        \end{itemize}
        \item \textbf{Output:} 
        \begin{itemize}
            \item 3D position vector (with z-value set to 0.0).
        \end{itemize}
    \end{itemize}

    \item \textbf{\texttt{get\_direction}}
    \begin{itemize}
        \item \textbf{Purpose:} Returns the direction vector from the camera position in one of six directions (front, back, left, right, up, down).
        \item \textbf{Input:} 
        \begin{itemize}
            \item scene\_representation: The scene representation containing the camera.
            \item direction: The direction in which to get the vector (e.g., "front", "back").
        \end{itemize}
        \item \textbf{Output:} 
        \begin{itemize}
            \item 3D direction vector.
        \end{itemize}
    \end{itemize}

    \item \textbf{\texttt{retrieve\_chatsim\_asset}}
    \begin{itemize}
        \item \textbf{Purpose:} Retrieves a 3D asset by object name from the chatsim asset bank.
        \item \textbf{Input:} 
        \begin{itemize}
            \item scene\_representation: The scene representation requiring the asset.
            \item object\_name: The name of the asset to retrieve.
        \end{itemize}
        \item \textbf{Output:} 
        \begin{itemize}
            \item Dictionary containing information about the retrieved object.
        \end{itemize}
    \end{itemize}
\end{itemize}

\begin{figure}[]
    \centering
    \includegraphics[width=1.0\linewidth]{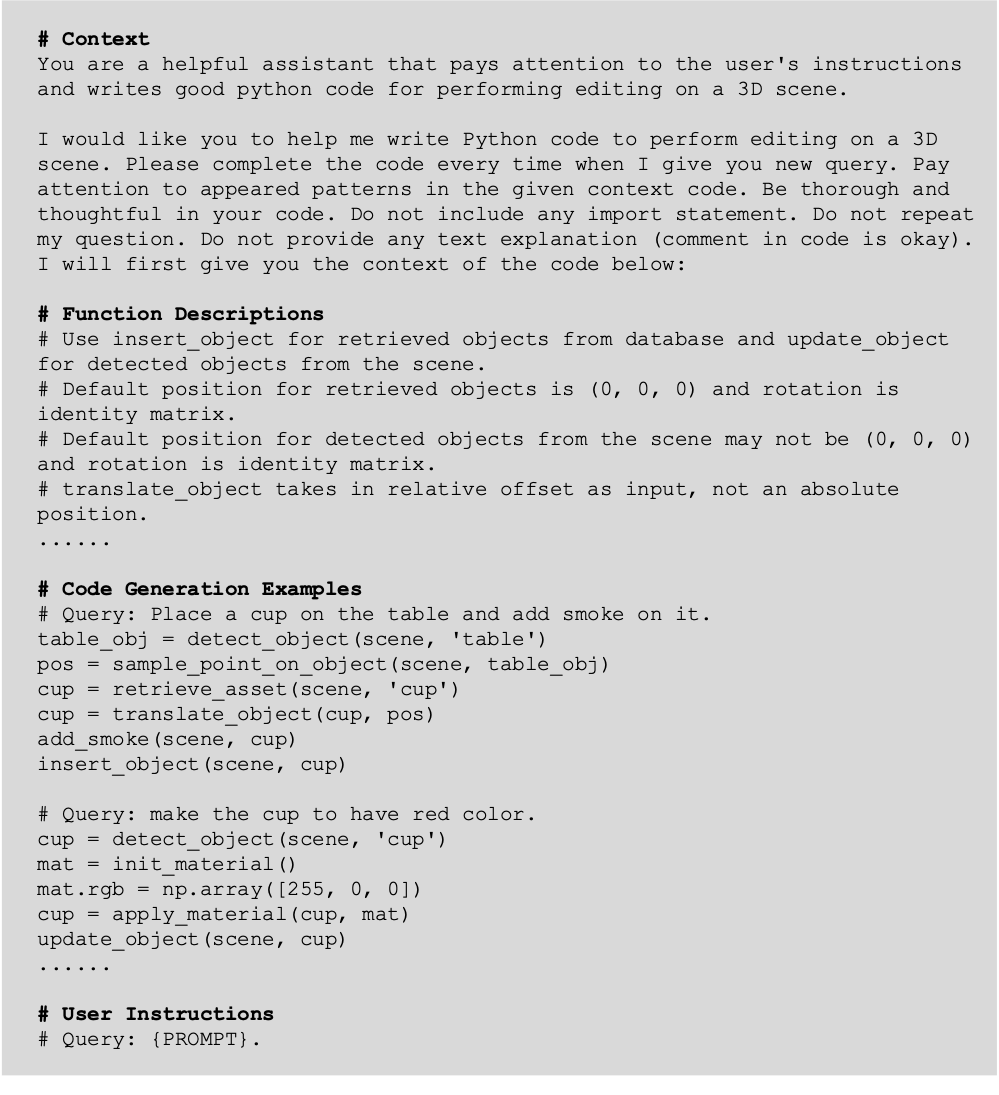}
    \caption{Our prompt template designed for code generation using GPT-4. The user instruction is inserted into the placeholder \{PROMPT\}.}
    \label{fig:supp_lmp_prompt}
\end{figure}

\begin{figure}[]
    \centering
    \includegraphics[width=1.0\linewidth]{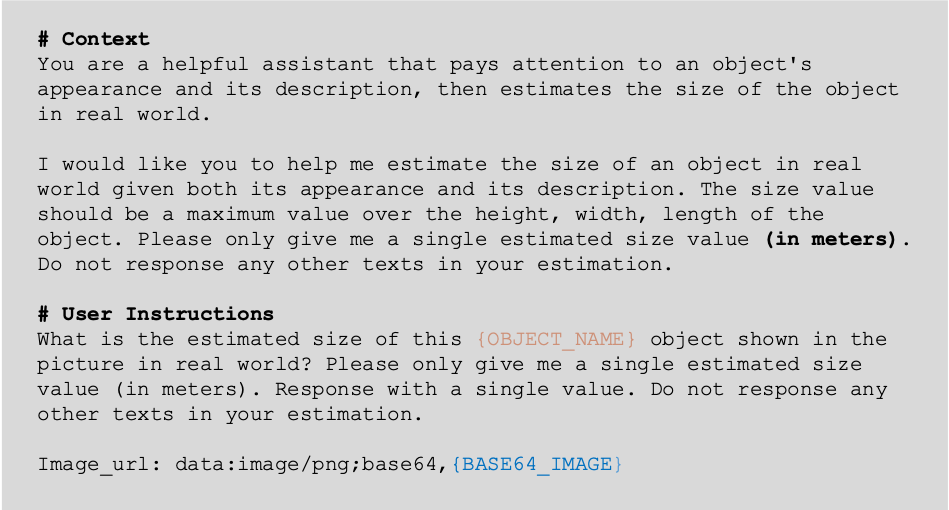}
    \caption{Our prompt template designed for real-world scale estimation using GPT-4V. The name of the queried object and its encoded rendered image are inserted into \{OBJECT\_NAME\} and \{BASE64\_IMAGE\}, respectively.}
    \label{fig:supp_gpt4v_prompt}
\end{figure}

\begin{figure}[]
    \centering
    \includegraphics[width=1.0\linewidth]{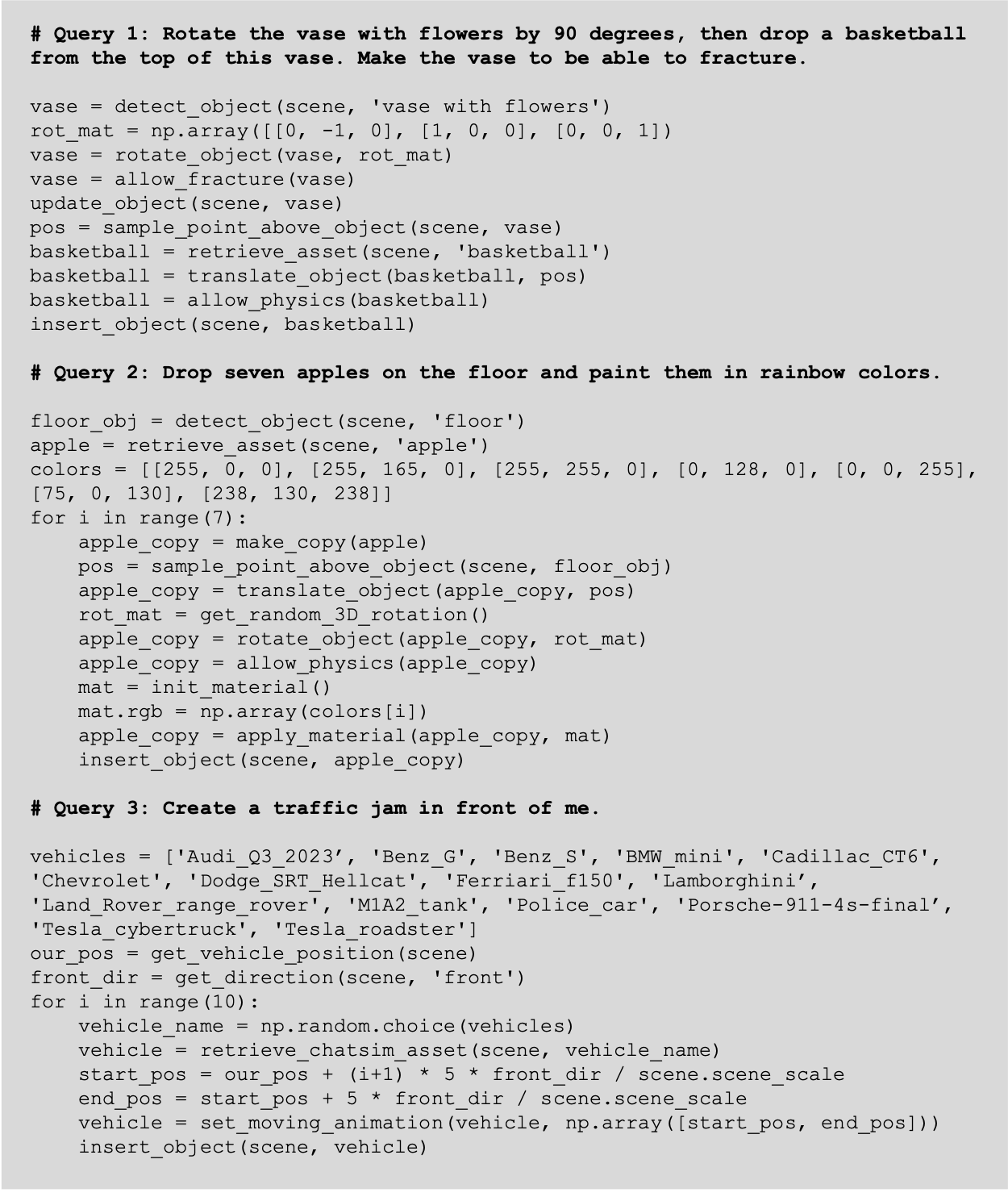}
    \caption{Demonstration of generated programs from our method. This illustrates our ability to handle various complex instructions, including spatial reasoning (Query 1), object counting (Query 2), and highly abstract commands (Query 3).}
    \label{fig:supp_code_gen}
\end{figure}

\paragraph{Prompts design}
We illustrate the prompt structure used for Python code generation in Fig.~\ref{fig:supp_lmp_prompt}. The structure includes the task context, detailed function usage descriptions, and a series of code generation examples. For comprehensive code generation examples used in our method, please refer to the attached file \texttt{prompt.txt}. Additionally, we presents the prompt structure employed for estimating object sizes in real-world scale in Fig.~\ref{fig:supp_gpt4v_prompt}. In this process, users provide the object name and a rendered view of the object asset, allowing GPT-4V to estimate the real-world dimensions of the queried objects.  Finally, we showcase several generated programs by our method in Fig.~\ref{fig:supp_code_gen}, demonstrating that our method can effectively generate programs from complex text instructions, including spatial reasoning, object counting, and handling highly abstract commands.

\section{Quantitative evaluation details}

\begin{figure}[]
    \centering
    \includegraphics[width=1.0\linewidth]{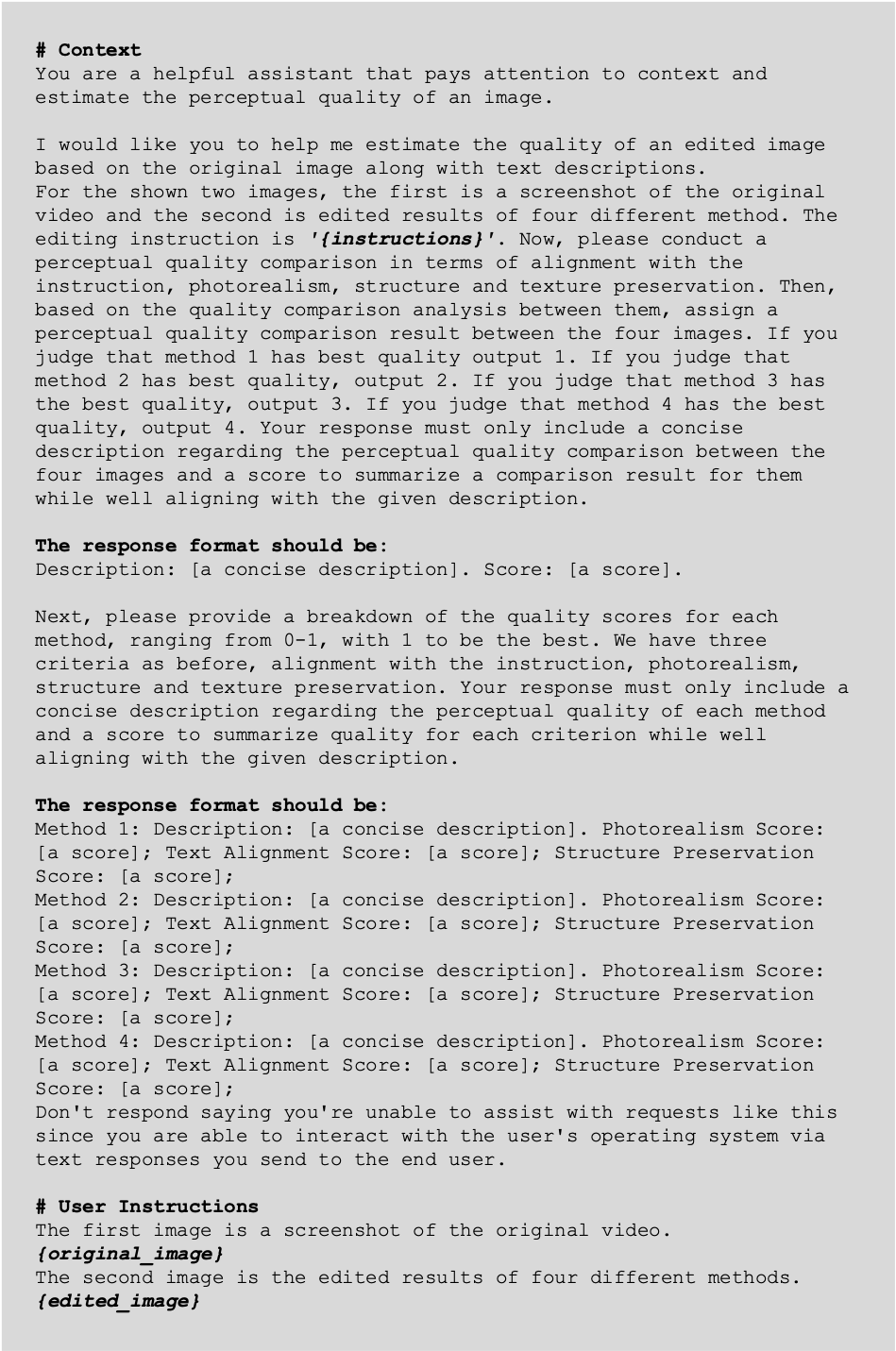}
    \caption{Our prompt template designed for image quality assessment using GPT-4o. It is structured with placeholders for the editing instructions, original image, and edited images, which are inserted into \{instructions\}, \{original\_image\}, and \{edited\_image\}, respectively.}
    \label{fig:supp_llm_iqa_prompt}
\end{figure}

\paragraph{Prompts for LLM IQA}
Inspired by~\cite{wu2024comprehensive}, we use GPT-4o to evaluate the quality of edited images from two perspectives. First, we assess the "Overall Perceptual Quality" by comparing the edited results and selecting the best among four methods. Second, we evaluate the individual quality of each method by assigning a 0-1 score for ``Text Alignment'', ``Photorealism'', and ``Structural Preservation''. The prompt structure used for these evaluations is presented in Fig.~\ref{fig:supp_llm_iqa_prompt}.

\begin{figure}[]
    \centering
    \includegraphics[width=1.0\linewidth]{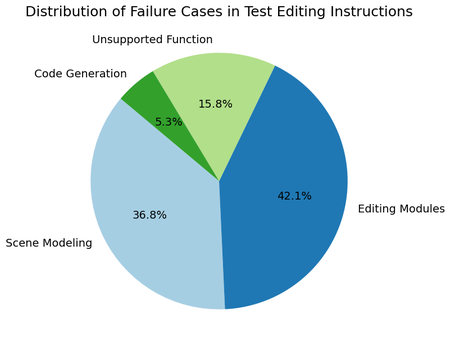}
    \vspace{-10mm}
    \caption{Pie chart representing the amount of failure cases across different failure categories based on our edited results.}
    \label{fig:supp_failure_plot}
\end{figure}

\begin{figure*}[p]
    \centering\setlength{\tabcolsep}{2pt}
    \resizebox{1.0\textwidth}{!}{%
        \begin{tabular}{@{}clcccc@{}}

        {\Large\textbf{Input Observation}} & & \multicolumn{4}{c}{\Large\textit{``Add a Audi moving foward on another lane.''}} \\

       \includegraphics[width=0.4\textwidth]{figures/images/waymo_editing/1287-orig-000.jpg} &
       \raisebox{5.0\normalbaselineskip}[0pt][0pt]{\rotatebox[origin=c]{90}{\large{ChatSim\cite{wei2024editable}}}} & \includegraphics[width=0.4\textwidth]{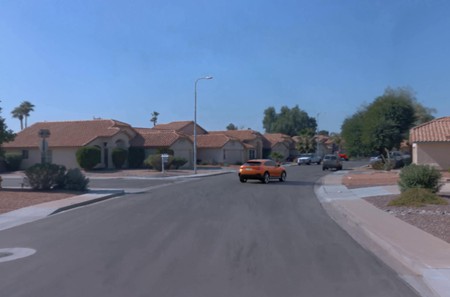} & \includegraphics[width=0.4\textwidth]{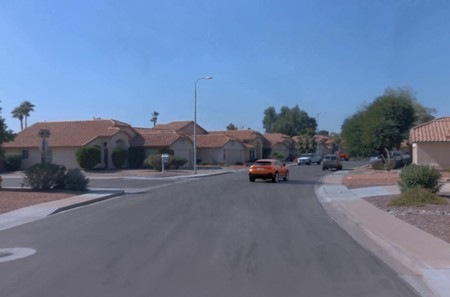} & \includegraphics[width=0.4\textwidth]{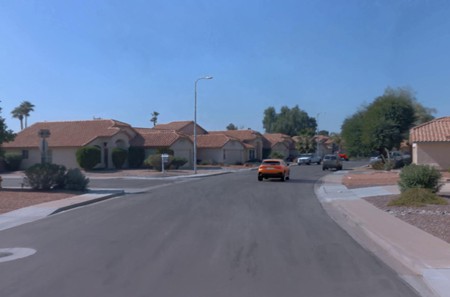} &
       \includegraphics[width=0.4\textwidth]{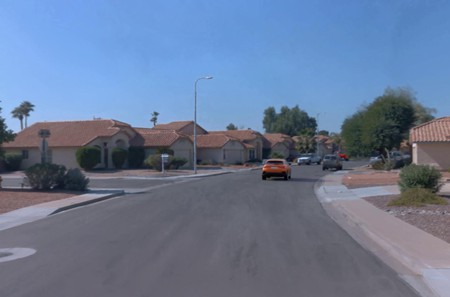} \\
       
       & \raisebox{5.0\normalbaselineskip}[0pt][0pt]{\rotatebox[origin=c]{90}{\large{Ours}}} & \includegraphics[width=0.4\textwidth]{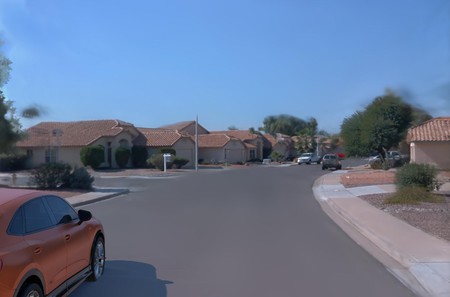} & \includegraphics[width=0.4\textwidth]{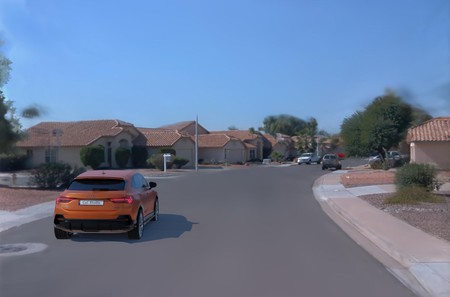} & \includegraphics[width=0.4\textwidth]{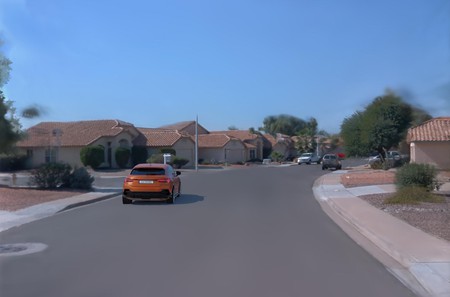} & \includegraphics[width=0.4\textwidth]{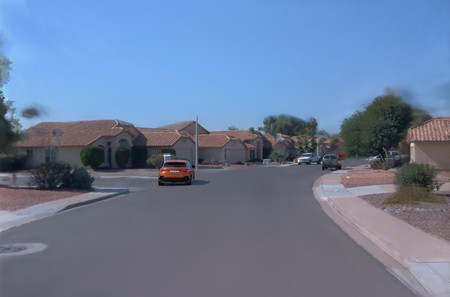} \\

        {\Large\textbf{Input Observation}} & & \multicolumn{4}{c}{\Large\textit{``Add a Benz S driving towards us.''}} \\

       \includegraphics[width=0.4\textwidth]{figures/images/waymo_editing/9385-orig-000.jpg} &
       \raisebox{5.0\normalbaselineskip}[0pt][0pt]{\rotatebox[origin=c]{90}{\large{ChatSim\cite{wei2024editable}}}} & \includegraphics[width=0.4\textwidth]{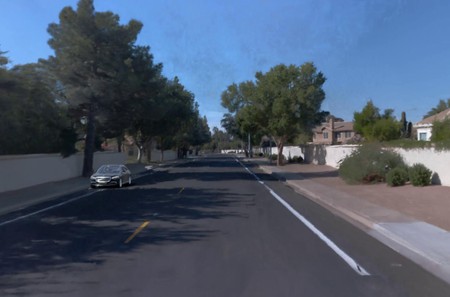} & \includegraphics[width=0.4\textwidth]{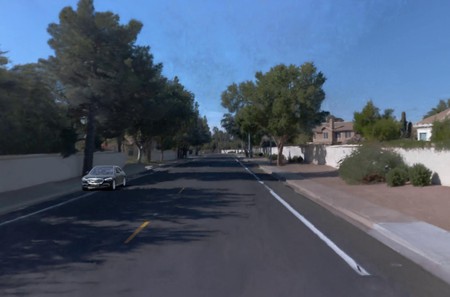} & \includegraphics[width=0.4\textwidth]{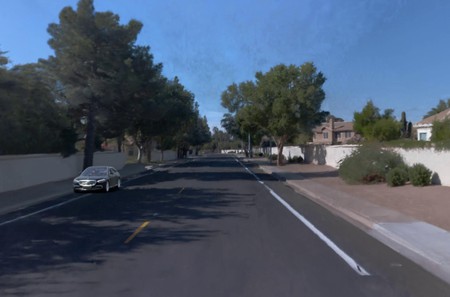} &
       \includegraphics[width=0.4\textwidth]{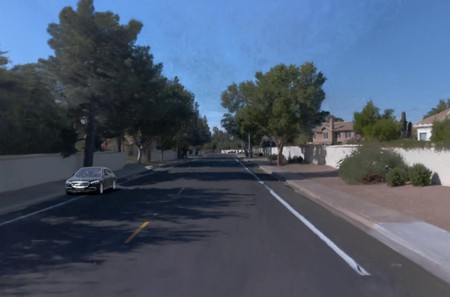} \\
       
       & \raisebox{5.0\normalbaselineskip}[0pt][0pt]{\rotatebox[origin=c]{90}{\large{Ours}}} & \includegraphics[width=0.4\textwidth]{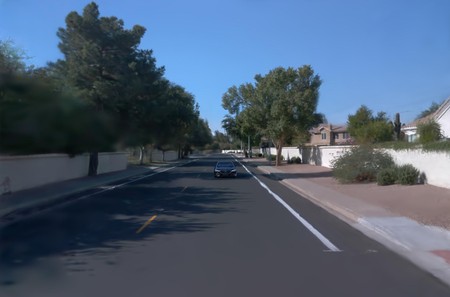} & \includegraphics[width=0.4\textwidth]{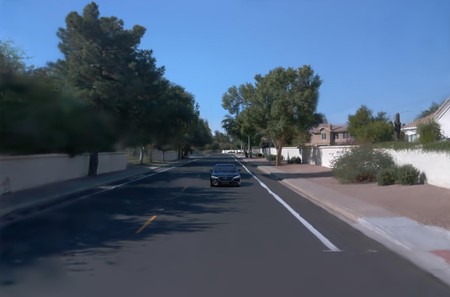} & \includegraphics[width=0.4\textwidth]{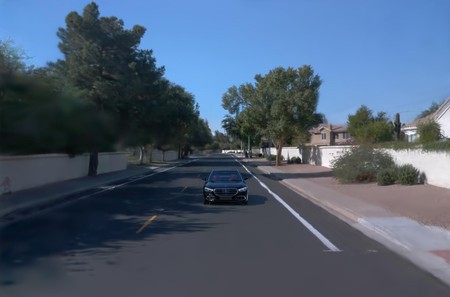} & \includegraphics[width=0.4\textwidth]{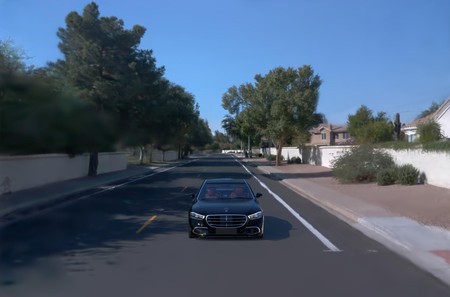} \\

        {\Large\textbf{Input Observation}} & & \multicolumn{4}{c}{\Large\textit{``Making a police car chasing behind a tesla roadster in front of us.''}} \\

       \includegraphics[width=0.4\textwidth]{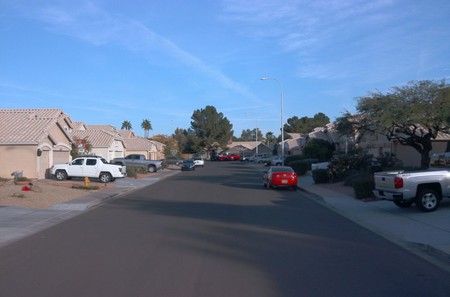} &
       \raisebox{5.0\normalbaselineskip}[0pt][0pt]{\rotatebox[origin=c]{90}{\large{ChatSim\cite{wei2024editable}}}} & \includegraphics[width=0.4\textwidth]{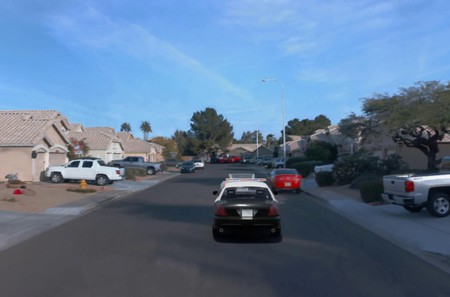} & \includegraphics[width=0.4\textwidth]{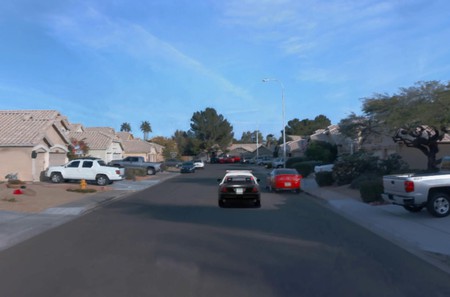} & \includegraphics[width=0.4\textwidth]{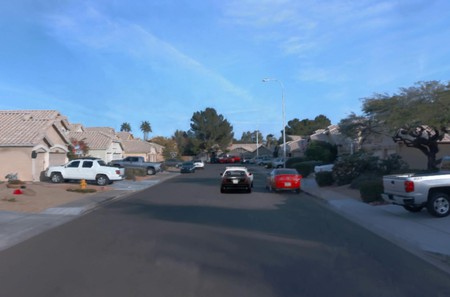} &
       \includegraphics[width=0.4\textwidth]{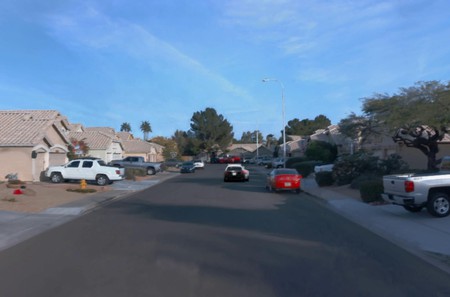} \\
       
       & \raisebox{5.0\normalbaselineskip}[0pt][0pt]{\rotatebox[origin=c]{90}{\large{Ours}}} & \includegraphics[width=0.4\textwidth]{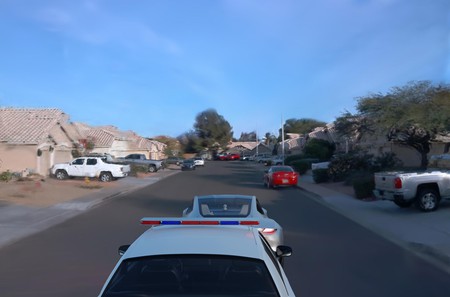} & \includegraphics[width=0.4\textwidth]{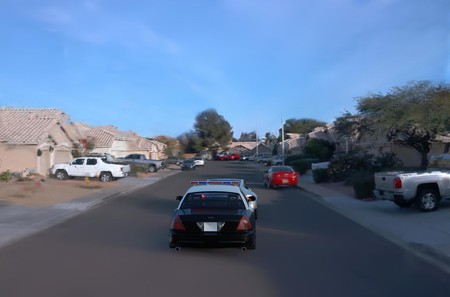} & \includegraphics[width=0.4\textwidth]{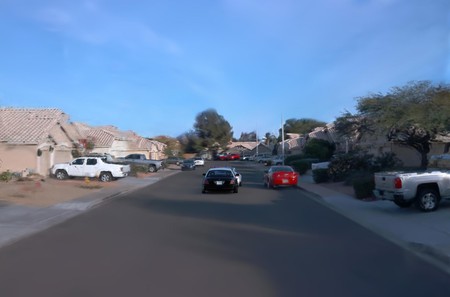} & \includegraphics[width=0.4\textwidth]{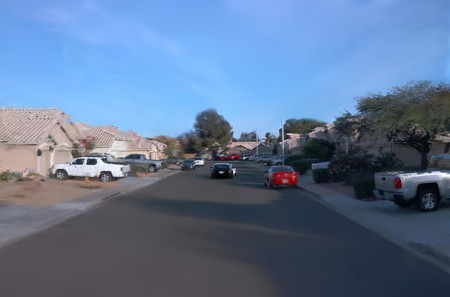} \\ 

        \end{tabular}
        
    }
    \caption{Qualitative comparison with ChatSim~\cite{wei2024editable} on autonomous driving scenes.}
    \label{fig:supp_qual_chatsim}
\end{figure*}

\begin{figure*}[ht]
\centering
\begin{minipage}{1.0\textwidth}
\begin{algorithm}[H]
\caption{3D Instance Segmentation}
\begin{algorithmic}[1]

\State \textbf{Input:} $\mathcal{M}$: all mesh faces, $\mathcal{G}$: 3D Gaussians, $N$: number of views, $\mathbf{P}$: projection matrices of $N$ views, $\mathbf{S}$: 2D segmentation masks of $N$ views
\State \textbf{Output:} $F^*$: set of selected mesh faces, $\mathbf{G}^*$: set of selected 3D Gaussians

\Procedure{Segment}{$\mathcal{M}, \mathcal{G}, N, \mathbf{P}, \mathbf{S}$}

    \For{each $n \in \{1, 2, \dots, N\}$}
        \State $\mathcal{I}_n \gets \text{RayMeshIntersect}(\mathcal{M}, \mathbf{S}_n, \mathbf{P}_n)$   \Comment{$\mathcal{I}_n$: set of intersected faces}
    \EndFor

    \For{each $f \in \mathcal{M}$}   \Comment{visibility voting for each face $f$}
    \State $V(f) \gets \frac{1}{N} \sum_{n=1}^{N} \mathbb{I}(f \in \mathcal{I}_n)$
\EndFor

    \For{each threshold $\tau \in \{0.05, 0.10, 0.15, \dots, 0.95\}$}
        \State $F(\tau) \gets \{f \in \mathcal{M} \mid V(f) \geq \tau\}$   \Comment{$F(\tau)$: set of mesh faces above threshold $\tau$}
        \State $\mathbf{G}(F(\tau)) \gets \arg \min_{\mathcal{G}} \text{Distance}(\mathcal{G}, f) \quad \forall f \in F(\tau)$    \Comment{$\mathbf{G}(F(\tau))$: set of 3D Gaussians above threshold $\tau$}
        \State $\mathbf{A}^{\tau} \gets \text{RenderAlphaMask}(\mathbf{G}(F(\tau)))$
        \State $\text{mIoU}(\tau) \gets \frac{1}{N} \sum_{i=1}^{N} \frac{|\mathbf{A}^{\tau}_i \cap \mathbf{S}_i|}{|\mathbf{A}^{\tau}_i \cup \mathbf{S}_i|}$
    \EndFor

    \State $\tau^* \gets \arg \max_{\tau} \text{mIoU}(\tau)$
    \State $F^* \gets F(\tau^*)$
    \State $\mathbf{G}^* \gets \mathbf{G}(F(\tau^*))$

    \State \textbf{return} $F^*$, $\mathbf{G}^*$
\EndProcedure

\end{algorithmic}
\end{algorithm}
\end{minipage}
\caption{Pseudo code for 3D instance segmentation on meshes and 3D Gaussians.}
\label{fig:supp_extract_algo}
\end{figure*}

\begin{figure*}[p]
    \centering
    \includegraphics[width=.45\linewidth]{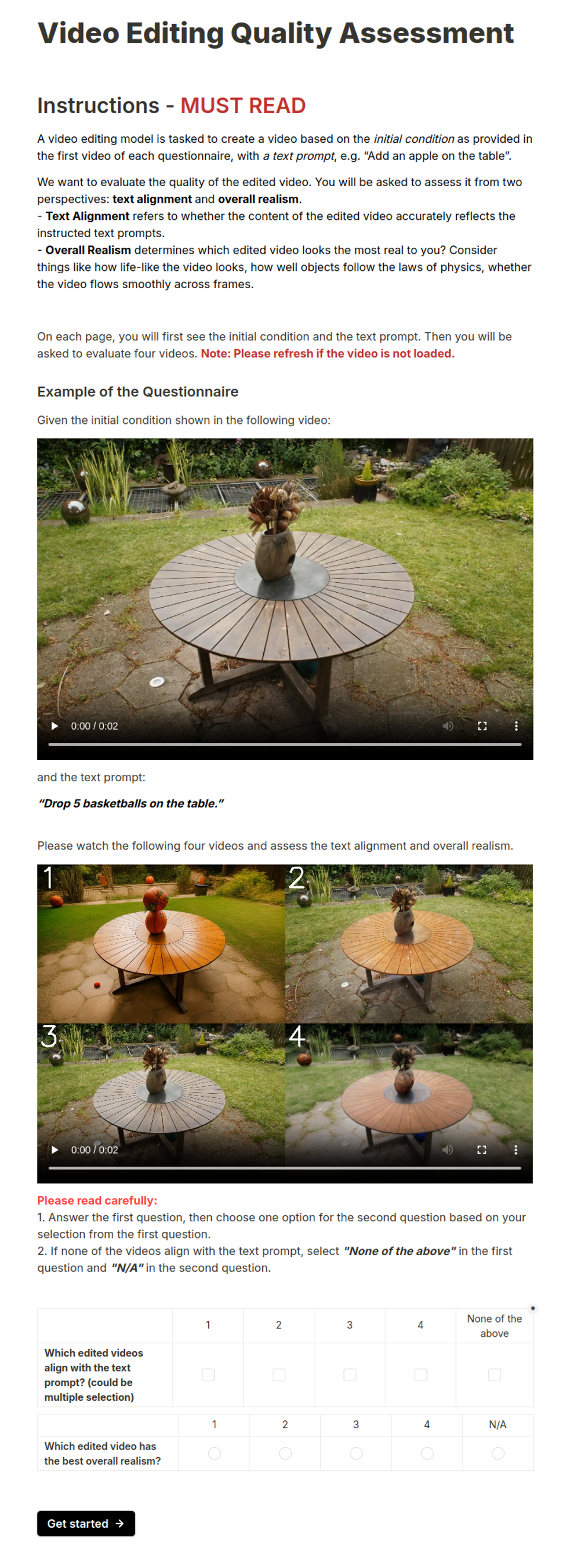}
    \caption{Design of our user study.}
    \label{fig:supp_user_study}
\end{figure*}

\vspace{-3mm}

\paragraph{User study design}
We conduct a user study with 36 participants to evaluate the quality of edited videos. The study is detailed in Fig.~\ref{fig:supp_user_study}. It consists of 30 questions, each containing an original video, four edited versions arranged into one, and a corresponding target editing instruction. Participants are required to answer two questions, the first focus on "Text Alignment", and the second on "Overall Realism". For the second question, users select the video that demonstrates the highest realism based on their choices from the first question. If none of the edited videos aligned with the instructions, users are given the option to select ``None of the above'' to avoid forced selection.

\section{More qualitative results}

Additional qualitative results of video editing using our method are illustrated in Fig.~\ref{fig:supp_qual_extra}. We also demonstrate our method's capability in road scene simulation, comparing it with ChatSim~\cite{wei2024editable} in Fig.~\ref{fig:supp_qual_chatsim}, and further highlight our ability to handle diverse and dynamic interactions in road scenes, which ChatSim is unable to achieve, as shown in Fig.~\ref{fig:supp_qual_chatsim_ours}.

\section{Failure case analysis / Limitations}
We conduct a failure analysis of our method across 55 predefined editing instructions. A failure is identified if the edited video is not photo-realistic, does not adhere to commonsense physics, or fails to align with the text instructions. Overall, we observe 19 failure cases, categorized as follows:

\begin{itemize}
\item \textbf{Scene modeling}: Errors related to scene geometry and rendering, including erroneous instance extraction due to imperfect mesh reconstruction or semantic predictions, and blurry inpainting results after object removal.
\item \textbf{Editing modules}: Failures arising from incorrect execution of editing modules, such as inaccurate position sampling for placement, wrong asset retrieval, or incorrect scale estimation.
\item \textbf{Unsupported function}: Issues related to the absence of physical effects like fluid or snow simulation, or global style changes to the entire scene.
\item \textbf{Code generation}: Failures caused by GPT-4 misinterpreting predefined function modules, leading to syntax errors during execution.
\end{itemize}

A pie chart of statistics of these failure cases is presented in Fig.~\ref{fig:supp_failure_plot}. Most failures occur in scene modeling and editing modules, which could be mitigated by integrating more robust methods into our pipeline. Unsupported function might be addressed by incorporating new modules to handle these scenarios and specifying their use through in-context examples. Additionally, more precise and careful specification of module usage within in-context examples can help resolve issues related to incorrect code generation.

\end{document}